\documentclass[12pt,leqno]{article}
\usepackage{times}
\usepackage{avm}
\usepackage{tipa}
\usepackage{amsmath}
\usepackage{amssymb}
\usepackage{natbib}
\usepackage{qtree}
\usepackage{pstricks}
\usepackage[utf8]{inputenc}
\usepackage[T1]{fontenc}
\usepackage{appendix}     
\bibpunct[:]{(}{)}{,}{a}{}{,}

\usepackage{graphicx}  

\usepackage{floatrow}  
\newfloatcommand{capbtabbox}{table}[][\FBwidth] 

\newtheorem{thrm}{Theorem}
\newtheorem{lem}{Lemma}

\usepackage{gb4e}
\def\glt{\nobreak\vskip.17\baselineskip}
\begin{document}

\begin{center}
{\Large\bf  The Emergence of Grammar through Reinforcement Learning} \\
Stephen Wechsler$^1$, James W. Shearer$^2$, and Katrin Erk$^1$ \\
$^1$\textit{The University of Texas at Austin}, $^2$\textit{JW Shearer Consulting}\\
{\tt wechsler@austin.utexas.edu}, {\tt JamesWShearer@gmail.com}, {\tt katrin.erk@utexas.edu }\\
\bigskip
\end{center}

\section{Complex grammar through learning}
\label{sec:introduction}

Human languages employ highly complex yet efficient grammatical systems for representing information about the world around us.  
Where does that efficient complexity come from?  
 This question is often addressed through the theory of learning, and that is the approach we take in this paper.  
However, addressing evolution and change through learning immediately presents us with a puzzle: if a child learns from witnessing the speech of others, then with perfect learning, her own speech may be expected to match that of her elders.  Then what drives grammar evolution?  Why do languages change?

We addressed this puzzle through \textsc{reinforcement learning} theory, drawing upon mathematical techniques in use in psychology since the 1950s \citep{bush1951math,bush1953stoch,bush1955stoch,
norman1968some,norman1972markov,harley1981learning,roth1995learning}.  
As it turns out, grammatical systems are predicted to emerge and grow in complexity,  
within our reinforcement learning models.  The crucial assumption that drives language change is that speakers have preferences for what messages to convey in a given context.  Those preferences  shape the development of grammatical systems, within our models.  Turning to human languages, it is clear that speakers in fact  have such preferences.  
  The thoughts and feelings that people choose to convey in an utterance depend upon their interests and goals, and on the role of the utterance itself as a social act woven into the text of human society.  Including the collective effects of those factors in our learning model makes it more realistic than it would be without them.  
Once we include these \textsc{message probabilities} in our model, a grammar can be shown to emerge, through the reinforcement learning theory alone.  In sum, we use a well-established learning theory to build a foundation for exploring the influential functionalist thesis that speakers' expressive purposes shape their language.  Then we use those models to derive languages in sufficient detail to test them against human languages.   

Our reinforcement learning rule has an alternative interpretation 
as a type of \textsc{frequency-based} learning, as favored in \textsc{usage-based} approaches to grammar emergence  \citep{hopper2001frequency,bybee2006usage,bybee2017}.   The likelihood that a speaker utters a given expression depends not only on the probability of their message, but also on how often and how recently they have heard the expression used in the past, and how often it was used to express the same or similar message.    
A speaker who considers uttering a phrasal sign uses their memories of that past input for two types of decision: deciding whether to use the sign, and, if they do, then also deciding on its syntactic form.  Message probabilities indirectly influence both decisions, in our models.  
Under many plausible conditions those effects are gradually amplified until speech production acquires the seemingly obligatory character of following conventions of syntactic and semantic composition.  

Specifically, in the simplest case a phrasal construction converges on the meaning of the most likely message in the given context (Section \ref{fundy-sec}).  However, we will show how system effects can cause some constructions to converge on a less likely meaning (Section \ref{similarity-sec}).   We will also see in detail how function morphemes become obligatory, and how complex rule systems emerge, such as a dependent case system (Section \ref{marked-sec}).  
Perhaps most interestingly, we will see the emergence of communicative efficiency:   languages are predicted to develop greater informativeness under conditions that systematically minimize the concomitant increase in formal complexity (Section \ref{efficiency-sec}).  This is an intriguing result in light of important studies demonstrating the efficiency of human languages (see \cite{Gibson+etal:2019} for an overview).

We introduce `forgetting', so that recent input data are more influential than input from the distant past  (Section \ref{forget-sec}). We observe that this improves the model in the sense that it speeds up the establishment of rules, especially in cases where the two most likely messages are close in probability.  Strikingly, the model is robust to cross-speaker variation in message biases:  the language of a heterogeneous speech community evolves in keeping with the \textit{average} message probability distribution of its members (Section \ref{variation-sec}).  

These predictions are not obvious \textit{a priori} and must be carefully demonstrated.  In this paper the demonstration takes various forms, including a set of theorems for which selected analytic proofs are provided  in the Appendix. Numerical simulations (Section \ref{numerical-sec}) illustrate the proofs and provide more
detail to the character of the learning trajectory and speed at which rules become established.

\section{Components of the model}
\label{components-sec}

\subsection{Reinforcement learning} 
\label{learning-sec}
This section provides background on theories of reinforcement learning, and situates our proposal relative to that background.
Reinforcement learning in psychology (as opposed to 
machine learning) refers to a family of mathematical models 
of how animals and humans learn.   
It has its origins with 
Thorndike's {\it Law of Effect}: 
behavior with positive outcomes is reinforced and likely to be
repeated (learned). 
Reinforcement learning is part of a larger family of 
stochastic learning models where behavior is probabilistic
\citep{bush1951math,bush1953stoch,bush1955stoch}.  
The key ideas are that the {\sc state of learning}
of a {\sc subject} (person or animal) is represented by 
a vector in a {\sc state space}. The subject's behavior 
(or {\sc response}) given a {\sc stimulus} 
is not deterministic, but depends on probabilities
determined by the state of learning. 
The {\sc outcome} (or {\sc payoff}) changes the state of learning. 
In reinforcement learning, the relative size of the payoff
determines how strongly (if at all) the behavior is reinforced. 
Over successive trials 
the state of learning 
changes along with behavior probabilities. 

The following simple  
reinforcement learning
model  
was first used in evolutionary game theory by 
 \cite{harley1981learning} and in economic game theory by
\cite{roth1995learning}.   
All of the learning formulas in this paper are based on the one in (\ref{HRE1}); we call them Harley-Roth-Erev (HRE) formulas.  
In this model, trial payoffs are 
non-negative numbers.  Previous learning models typically assume $I$ different behaviors for the learner
  to choose from; the state of learning vector
at the $k$th trial is 
${\bold c} = (c_1, c_2, \ldots, c_I)$, where $c_i$ is the cumulative
sum of payoffs for the $i$th behavior after $k-1$
trials. 
The probability $p(i|c)$ of choosing the $i$th behavior
at the $k$th trial with state of learning \textbf{c} is given by (\ref{HRE1}):

\begin{exe}\ex\label{HRE1} Harley-Roth-Erev (HRE) formula
\begin{align*} 
p(i|\textbf{c})
&\ =\
\frac{ c_i }{ c_1 + \ldots + c_I }
\end{align*}
\end{exe}
The behavior with highest cumulative sum of payoffs has the highest
probability of being chosen. Higher and more frequent
payoffs make that behavior
more likely. 
For the first trial $k=1$, one must choose 
an initial state of learning vector that can incorporate 
prior assumptions of behavior probabilities.  
Roth and Erev use the term {\sc propensities} for the 
cumulative sums $c_i$ ($c_1$, $c_2$, \dots) in an HRE equation, and we occasionally use that term below. 

In our application to language learning, the HRE formula gives the probability that the learner produces a specified target 
utterance type $u^\star$, given a message $m_i$ that she seeks to convey.  
But  the learner does not use the HRE formula to choose among $I$ different behaviors, as she did in the previous models described above.  Instead, the learner uses it for a simple binary choice between uttering and refraining from uttering a single target sign $u^\star$ (see the Fundamental Model, Sec. \ref{fundy-sec}).  
The propensities are determined by counts ($c_1, c_2, \ldots, c_I$) of previous utterances  that the speaker has witnessed, but those counts do not correspond to different possible behaviors, such as different utterance forms.  Rather, the counts are all of $u^\star$ utterances, indexed by the meanings ($m_1, m_2, \ldots, m_I$) they conveyed  in the past.  Intuitively, the speaker can be seen as using the HRE formula to estimate the probability that $u^\star$ conveys her desired message $m_i$.  

In our Sequential Model (Sec. \ref{multiple-sec}), a speaker who refrains from uttering $u^\star$ starts over with a different target utterance form to express her message.  This continues through a sequence of utterance forms until she settles on a form to utter.  
When the speaker with message $m_i$ utters $u^\star$,  the  count $c_i$ is increased for the next learning trial, while the other counts ($c_j, j \neq i$) are not.   
Note that in (\ref{HRE1}),
$c_i$ appears as the numerator and is also an addend in the denominator.  Hence each time a speaker with message $m_i$ produces a $u^\star$ utterance, it increases the probability that the next speaker with message $m_i$, will also produce a $u^\star$ utterance.

When is learning complete, within the learning model?  
The $i$th behavior is fully learned
if the probability of $i$th behavior $p(i)$  \textsc{converges} to 1 as the number of trials $k$ grows large.\footnote{Convergence is in the probabilistic sense of `almost sure'.  See the theorem proofs in the Appendix.}  
Intuitively convergence to 1 means that a speaker with the given message in mind is theoretically predicted to produce the utterance $u_i$.  There are other types of learning that lead to a random mix of behaviors
and converge to a limit probability distribution over multiple behaviors.  (We look at such `free variation' in Section \ref{marked-freq-sec}.)  In this paper we study the convergence properties of learning models under different conditions.  We show that syntax emerges under all plausible conditions and that many plausible conditions  give rise to syntactic systems resembling those of human languages.

The various types of convergence along with its rate can usually be
observed in numerical simulations and sometimes 
proven as a theorem. 
In the 1960's \cite{norman1968some} 
developed the mathematics behind
stochastic learning models, interpreted them as Markov processes and
proved some general convergence theorems \citep{norman1972markov}.
 \cite{beggs2005convergence} proved convergence
theorems for some models in \citet{roth1995learning}.
Our models are Markov processes and our theorems and 
numerical results will be concerned with convergence and its
rate. We use results in Norman, Beggs and stochastic 
approximation theory in \cite{Pemantle2007} and \citet{benaim2006} 
in some of our proofs.  

The reinforcement learning rule HRE has many nice properties.
HRE is consistent with two well documented features of 
human and animal learning. 
The first is Thorndike's \textit{Law of Effect} noted above and the second is
the \textit{Power Law of 
Practice}, where learning is initially fast, then slows down.  
As noted in the introduction above, HRE has an alternative interpretation 
as a type of frequency based learning.
 If the payoffs
are 1 for success and 0 for failure, then the state
of learning $c_i$ equals 
the count of instances of the $i$th behavior.  In linguistic research this can be operationalized by counting tokens of the $i$th behavior in a corpus, as we do in our case studies.  
Moreover, HRE has relatively low cognitive demands, yet, as shown both in 
 \cite{harley1981learning} and in
\cite{roth1995learning}, one can often 
quickly learn an approximate optimal strategy 
in the sense that HRE gets close to 
an {\it evolutionarily stable strategy} (ESS) or a Nash equilibrium
in evolutionary or economic games. 
In contrast,
finding ESSs or Nash equilibria is typically cognitively expensive,
even for simple games. Hence, this supports Harley's 
thesis that animals and humans have evolved
relatively simple 
learning rules that (i) would generally find good strategies quickly
and (ii) have the same asymptotic properties as HRE.    
Lastly, both  \cite{harley1981learning} and 
\cite{roth1995learning} found simulated HRE learning often
tracked well with empirical results and observations 
of animal and human learning.

For several reasons we may want a model where initial 
learning starts very slowly, and this will be done with an 
additional parameter $\alpha \geq 0$ in the denominator on an HRE formula:  
\begin{exe}\ex\label{HRE2}  HRE formula with parameter $\alpha$
\begin{align*} 
p(i|\textbf{c})
&\ =\
\frac{ c_i }{ c_1 + \ldots + c_I + \alpha}
\end{align*}
\end{exe}
In the case of a positive value for $\alpha$, we are assuming  
most outcomes in trials initially have \textsc{ineffective conditioning}
where the state of learning is unchanged. And then eventually 
most trials lead to \textsc{effective conditioning} with changes to
the state and the probabilities of behaviors.  

\

\subsection{Model parameters from human cognition}

We posit two types of background condition, \textsc{message probabilities} and \textsc{form probabilities}.  In addition, learning that involves imitation, including language learning, is influenced by the learner's \textsc{similarity judgments}.  

\bigskip
\noindent 
\textbf{Message probabilities.}  Language gives us apparently infinite expressive capability, but the patterns of daily life lead to patterns of preference for what to express.  In the models presented below, different messages can be more or less likely in a given type of utterance, within a given context.   
Those preferences are influenced by innate and environmental factors, but the exact origins of the probabilities will not concern us here.  In fact one striking result, reported in Sec. \ref{variation-sec} below, is that homogeneous linguistic conventions are predicted to emerge even within a speech community whose members have diverse interests and thus varied message probability distributions.   The emergent linguistic  conventions are predicted to reflect the \textit{average} message probability distribution of the speech community.  

Message probabilities play an important role in all of the models below.  But they play themselves, so to speak.  We do not posit an extrinsic causal link to the grammar, but rather derive the effects of their presence directly.  Message probabilities interact with the learning model to influence language evolution in two ways: in the Fundamental Model (Sec. \ref{fundy-sec}) they drive the emergence of grammatical relations by biasing the selection of a semantic composition rule; and they lead to a dissimilation between formal expressions of distinct grammatical relations (Sec. \ref{cat-sec}).  

\bigskip
\noindent
\textbf{Form probabilities.}    
The forms that express the emergent grammatical relations are also subject to biases.  Form biases have been studied extensively with an eye to understanding their origins and the reasons for typological variation.  We illustrate below with a subject-first rule that emerges due to an `easy-first' processing bias (See Sec. \ref{form-sec}). 

\bigskip
\noindent
\textbf{Similarity judgments.}  We learn how to use a verb in a sentence based on exposure to prior utterances of sentences containing the verb.  What if there are no prior sentences containing the verb?  In the Model with Similarity (Sec. \ref{similarity-sec}), speakers count not only sentences with the verb but also those with semantically \textsc{similar} verbs.  To do that, they must judge similarity of semantic roles across verbs:  for example, running is similar to walking.   (See Section \ref{similarity-sec}.) 

\bigskip
The three classes of assumptions given above seem to be uncontroversial, at least at a general level.  With the exception of the case studies (Sec. \ref{dia-sec}), we don't attempt to motivate specific probabilities or specific similarity measures.  Instead our strategy will be to show that grammar emerges under all parameter settings, no matter how implausibly pessimistic, and that grammar emerges reasonably quickly under all plausible settings.

\subsection{Relation to other proposals}
\label{litreview-sec}

The models presented below build on a rich and growing tradition of mathematical modeling of language evolution.  See, among others, \citet{kirby1998language}, the  papers in \citet{briscoe2002book},  
\citet{skyrms2010signals,kirby+etal2015,huttegger+etal:2014}; and see especially \cite{spike2017minimal} for insightful discussion and comparison of various approaches, including those of \citet{nowak+Krakauer:1999,steels2012self,barrett2006numerical,franke2012bidirectional,Oliphant+Batali:1997,smith2002cultural} and \citet{barr2004establishing}.  
Many proposals involve some form of reinforcement learning: e.g. agents in \citet{skyrms2010signals} use  HRE formulas to select a signal in a Lewis signaling game, and \citet[635]{spike2017minimal} provide an equivalent formula (their equation (1), p. 635) which an agent uses to select a signal from a list of options.  While our models are similar, they nonetheless differ in that our speakers use the HRE formula for a sequence of binary choices, as noted in the paragraph following (\ref{HRE1}) above.  However, a detailed comparison with other proposals will have to await later work, due to a lack of space.  For now we shall strive to present our assumptions, the scope of our work, and our results as clearly as possible.

\section{The Fundamental Model}
\label{fundy-sec}
Our first model of the emergence of syntax 
is called the Fundamental Model.  
It is `fundamental' in that it provides the essential foundation for syntax, on which more will be built in later sections of the paper.    

\subsection{Language histories}
A language history ($\mathcal{LH}$)  is modeled as a sequence of \textit{utterances}:

\begin{exe}\ex\label{lh}
$\mathcal{LH}$ = $\langle u^1, u^2, \ldots  \rangle$
\end{exe}
Each utterance ($u^k$) consists of a \textit{message} ($m$), an \textit{act of reference} ($r$) modeled as a map from a structured sentence (`forms', $f$) onto a structured \textit{scene} ($s$), and an utterance index $k$ whose values are shown as superscripts in (\ref{lh}) (in what follows these superscripts will often be suppressed when unneeded):

\begin{exe}\ex\label{map}
$u^k = \langle k,  m^k, r^k:f^k \rightarrow s^k \rangle$
\end{exe}
Figure 1 shows two acts of reference, each one in a 1-word utterance, to scenes of a cat walking in the grass.  Figure \ref{2word-fig} shows two acts of reference,  each one in a 2-word phrasal utterance, to scenes of a cat walking in the grass.  

\begin{figure}[h] 
$r_1$ =  {\avmoptions{center}
\begin{avm}
\[ \textsc{form}  & $\langle$  \textit{Cat.}$_{\@1} \rangle$\\  \textsc{scene} &  s\[ {\sc event} & walking \\ \textsc{walker} & {\@1}cat \\ \textsc{surface} & grass\] 
\] \ \ $r_2$ =  
\[ \textsc{form}  & $\langle$  \textit{Walk.}$_{\@1}   \rangle$\\  
\textsc{scene} &  s\[ {\sc event} &  {\@1}walking \\ 
\textsc{walker} & cat \\ \textsc{surface} &  grass\] 
\] 
\end{avm}
}
\caption{Acts of reference (form-scene maps) from two sample 1-word utterances }
\label{1word-fig}
\end{figure}

\begin{figure}[h]
$r_3$ =  {\avmoptions{center}
\begin{avm}
\[ \textsc{form}  & $\langle$  \textit{Cat}$_{\@1}$ \textit{walk}$_{\@2}. \rangle$\\  \textsc{scene} &  s\[ {\sc event} &  {\@2}walking \\ 
\textsc{walker} & {\@1}cat \\ \textsc{surface} & grass\] 
\] \ \ $r_4$ =  
\[ \textsc{form}  & $\langle$  \textit{Grass}$_{\@1}$ \textit{walk}$_{\@2}. \rangle$\\  
\textsc{scene} &  s\[ {\sc event} &  {\@2}walking \\ \textsc{walker} & cat \\ \textsc{surface} & {\@1}grass\] 
\]
\end{avm}
}
\caption{Acts of reference (form-scene maps) from two sample 2-word phrasal utterances}
\label{2word-fig}
\end{figure}

We make certain simplifying assumptions.  
All members of the speech community witness every utterance, hence we ignore network effects such as diffusion, language contact and dialect formation.  Speakers are also learners, with no distinction drawn between children and adults.  In the first model, speakers are immortal and remember long-past and recent input equally well, 
but in Section \ref{forget-sec} we introduce forgetting
by down-weighting 
past history as a way to model the effects of
memory and mortality.  

\subsection{Reinforcement learning with message probabilities}

We demonstrate grammar emergence through reinforcement learning with a simple model we call  Cat Walking in Grass.  
Over and over again, speakers report on  scenes they witness, of a cat walking in the grass; call this type of scene $s_{\textsc{walk}}$.  The lexicon has three words (sound-meaning pairs): \textit{cat}, \textit{walk}, and \textit{grass}.  The speaker's message always calls attention to the walking event, and therefore they always say the word \textit{walk}; and the speaker makes note of either the walker (by saying \textit{cat}) or the walking surface (by saying \textit{grass}).  Thus the speaker can say two words independently without syntax: `Cat. Walk.'; or `Grass. Walk.'  But since signs are sound-meaning pairs, we will assume a non-zero probability that the speaker conveys their meaning with a single complex sign, either [Cat walk.] or [Grass walk.], depending on their 
message.  \footnote{We use subject-verb order for simplicity, to represent a phrasal utterance expressing the relevant message.  Word order and other aspects of grammatical form are discussed later. 
For now the options we consider are whether to 
to utter a phrasal sign ($u^\star$), or not ($u^\dagger$).}  So this model has four utterance forms altogether.

A language history produced by the Cat Walking in Grass model is a sequence of utterances of exactly those four kinds:

\bigskip

\begin{exe}\ex \label{cwg-table}
A language history\\  
 \smallskip  
\begin{tabular}{ clcc } 
 \hline
$k$  & form & message & utterance \\ 
 \hline
1. & Cat. Walk. & m_1 & $u_1^\dagger$\\ 
2. & Cat walk. & m_1 & $u_1^\star$\\ 
3. & Grass. Walk. & m_2 & $u_2^\dagger$\\ 
4. & Grass walk. & m_2 & $u_2^\star$\\ 
5. & Cat walk. & m_1 & $u_1^\star$\\ 
6. & etc. & \ldots  & \ldots \\ 
 \hline
\end{tabular}
\end{exe}
Here $m_1 = m_{\textsc{walker}}$  represents the message `A cat is walking' and $m_2 = m_{\textsc{surface}}$  represents `Grass is being walked on'.  
Since we are modeling the emergence of phrases, a phrasal utterance is termed a \textsc{success}, indicated by the star in $u^\star$.   The dagger in $u^\dagger$ indicates \textsc{failure}: the speaker fails to make a phrase, and instead utters two separate signs.

A scene viewed by a speaker 
has various features of greater or lesser noteworthiness
to the speaker.  We model this by saying that
given a scene $s$, the speaker has 
a finite set of messages $m_1, m_2, \ldots, m_I$
with probabilities $p(m_i | s)$ summing to 1 ($\sum_i p(m_i | s) = 1$).
And $p(m_i | s )$ is the probability the speaker 
intends message $m_i$, given scene $s$.  
In the Cat Walking in Grass model $m_1$ and $m_2$  are the only two
message types observed for $s_{\textsc{walk}}$, so their probabilities sum up to 
one:\footnote{Message
probabilities are constant across scenes of a given type, such as $s_{\textsc{walk}}$, in the Fundamental  
Model, e.g. $p(m_i | s ) = p(m_i | s' ) $ for all $s$, $s'$, cat walking 
on grass scenes.  In the General Model (Section \ref{variation-sec}) we allow them to vary.  Also, in the Fundamental Model all members of the speech community (speakers and hearers) have the same message probabilities, while in the General Model this is not assumed.}

\begin{exe} \ex \label{walkprob1}
$1
= 
p(m_1 | s_{\textsc{walk}} ) +  
p(m_2  | s_{\textsc{walk}}  )  $
\end{exe}  
Given their message, the speaker then chooses to utter a phrasal sign (success, $u^\star$) or not (failure, $u^\dagger$).  
The probability of success for message $m_i$, scene $s$, and state of learning \textbf{c} is notated as $p( u^\star |  m_i, s, c)$.  
Its value is given by an HRE
equation, 
where it is determined by the counts $(c_1, c_2)$ of previously witnessed utterances.  We are suppressing in (\ref{walkHREs}) the dependence on scene
$s$ and state of learning \textbf{c}.  
We include a constant
parameter $\alpha \geq 0$ in the denominator.  A positive value for $\alpha$ depresses the overall likelihood of phrasal syntax emerging, as discussed below.  

\begin{exe}  \ex \label{walkHREs} Harley-Roth-Erev formulas for the Cat Walking in Grass model 
\end{exe}   

\begin{center}
$p(u_1^\star | m_1 )
=
\dfrac{ c_1 }{ c_1 + c_2 + \alpha}$
\ \ \ \ \ 
\ \ \ \ \ 
\ \ \ \ \ 
$p(u_2^\star | m_2 )
=
\dfrac{ c_2 }{ c_1 +  c_2 + \alpha}$
\end{center}

\bigskip

\noindent
In equations (\ref{walkHREs}), $c_1$ and $c_2$ are the counts of previous phrasal signs expressing messages $m_1$ and $m_2$, respectively.  
If the outcome is a success then the state of learning (including the counts $c_1$ and $c_2$) is updated accordingly, for the sake of the next utterance attempt.  

Let us review the roles of the various elements of the model 
in the terminology of reinforcement learning theory.  
The \textsc{stimulus} is the scene-message pair $(s,m_i)$
and the \textsc{response} is the attempted utterance. 
A successful phrasal utterance ($u^\star$ expressing $m_i$) acts as \textsc{effective conditioning} for language learning; it influences the learning state for the next attempt and thereby contributes to future speakers' likelihood of uttering phrase $u^\star$ to express $m_i$.  The values  that affect learning, such as the utterance counts $c_1$ and $c_2$, are called \textsc{propensities}.   The amount added to $c_1$ or $c_2$ for each phrasal utterance in the language history is called the utterance's \textsc{payoff size}; in the production algorithm just below, the payoff size is set at 1.

\subsection{A production algorithm for the Cat Walking in Grass model}
\label{prod-sec}

This production algorithm produces the $k$th utterance ($u^k$) in a language history.
  

\textbf{Step 1. Select a message.}  
Given a walking scene $s^k_{\textsc{walk}}$ of a cat walking in the grass,  
the speaker selects a message, either $m_1$ or $m_2$, from the probability distribution in (\ref{walkprob1}).

\textbf{Step 2. Produce an utterance.} The speaker decides whether to express her message as a two-word phrase ($u^\star$)  using the HRE probabilities in (\ref{walkHREs}), where: 
$i$ ranges over 1 and 2; $m_1 = m_{\textsc{walker}}$  and $m_2 = m_{\textsc{surface}}$;  
 $u_i^\star$ is a phrase expressing message type $m_i$; and  
 $c_i$ is the sum of a positive starting 
value  
$c_i^0$ plus the number of phrases $u_i^\star$
among the previous $k-1$ utterances.\footnote{\label{start-fn} Positive starting values  
are necessary, as negative values don't occur in a positive reinforcement model and 
if $c_i^0 = 0$ then 
equation (\ref{walkHREs}) would always start and stay at 0 since 
phrasal utterances would never occur.}

If the speaker does not utter $u^{\star}_i$ then they utter the same two words with no syntactic relation, 
which we notate $u^\dagger$.
Hence $p(u^\dagger | m_i) = 1 - p(u^{\star} | m_i)$.

\textbf{Step 3. Update the history.} Update the phrasal utterance counts $c_1$ and $c_2$ to reflect the outcome in Step 2 and return to Step 1.
More precisely, keep  $c_1$ and $c_2$ 
unchanged unless the utterance in 
Step 2 is a phrase $u_i^\star$.  In that case,
add 1 to $c_i$ and then return to Step 1. 

\begin{exe}\ex \label{step3}  Updating for utterance $u_i^\star$ (with message $m_i$):
\medskip \\
$c^k_i  = c^{k-1}_i + 1  
\medskip \\
c^k_j  =  c^{k-1}_j , $ where $ j\neq i $
\end{exe}
This process generates one random language history.  

\subsection{A fundamental result: the emergence of semantic composition}  
\label{theorem-sec}
We have investigated language histories generated by the Cat Walking in Grass model using both numerical simulations and analytical techniques, and we report on the results in this section.  These results are important to understand, as they form the basis for our theory of the emergence of semantic composition.  

Two Harley-Roth-Erev rules are used
in the production algorithm above, one for each message.
They provide the changing value of
$p(u^{\star}_i| m_i)$ over the course of a language history.  If
$p(u^{\star}_i| m_i)$ converges to 1 then $u_i$ is a grammatical
phrase of the language and it expresses $m_i$; if $p(u^{\star}_i|
m_i)$ converges to 0 then $u_i$ is not grammatical as an expression of
$m_i$.  What we have found, using both analytical techniques and numerical simulations, is that in every language history, 
for all starting values of $c_i^0>0$ (see footnote \ref{start-fn}) and any 
$\alpha \geq 0$,
one of
the two probabilities, $p(u^{\star}_1| m_1)$ or $p(u^{\star}_2| m_2)$,
converges to 1 and the other converges to 0.   
Specifically, the
utterance that converges to 1 expresses the message with the higher
probability in the distribution in
(\ref{walkprob1}).\footnote{If the two highest probabilities are exactly equal then neither one converges to 1, in the Fundamental Model.   We add forgetting to the model to get 
convergence even when the two message probabilities
are equal (see Section \ref{forget-sec}).} 
Assuming that speakers are more likely overall to mention the walker than the surface, then [\textit{Cat walk.}] becomes a conventional grammatical phrase while [\textit{Grass walk}.] does not.   As $k$ gets large, the language history shown in (\ref{cwg-table}) converges on two utterance forms, [\textit{Cat walk.}] and [\textit{Grass. Walk.}], while [\textit{Grass walk.}] and [\textit{Cat. Walk.}] disappear from the language.  

To state this result in more general terms we define
\textsc{convergence of a language history} as follows:
\begin{exe}\ex 
{\bf Defn:} a language history \textsc{converges} 
if each Harley-Roth-Erev rule converges to 1 or 0.\\
In that case, exactly one Harley-Roth-Erev rule
 converges to 1, others converge to 0.
\end{exe}
Then we can say that in the Cat Walking in Grass model, every language history converges, for any starting values of $c_i^0>0$ (see footnote \ref{start-fn}) and any value of the parameter $\alpha \geq 0$. 

We used two different methods in order to understand both qualitative and highly probable properties 
of a randomly generated language history.  First,
we used analytic methods to  precisely state those properties and prove an important general theorem, the \textit{Fundamental Theorem: Emergence of Semantic Composition}.  Second, we  numerically 
computed many language histories (sample paths), 
in order to observe in a 
probabilistic sense both the emergence of syntax
and the speed at which it happens.     
We have used such numerical methods to investigate 
qualitative properties of the language history, 
such as slow or fast emergence of
syntax.  The general theorem is presented next.  
The numerical simulations are discussed in Section \ref{numerical-sec}.

Why does it work?  The speaker must decide whether the phrasal sign expresses their intended message.  If they intend $m_1$ then they decide based on the counts $c_1$ and $c_2$.   If the speaker utters the phrasal sign, then $c_1$ grows by one, making that utterance a little more likely for the next speaker.  All the same is true for $m_2$ and $c_2$, of course.  But there are more opportunities for $c_1$ to grow, because $m_1$ is the favorite message overall.  So $c_1$ grows faster.  More precisely, 
the following formula gives the probability 
the $k$th  utterance mapping to scene $s$
 is a phrase ($u^\star$). 
\begin{exe}\ex \label{chain}
$p(u_i^{\star} | s)
=
p(u_i | m_i, s)
p( m_i | s)$
\end{exe}
(The state of learning $c^k$ is suppressed in (\ref{chain}).) If a speaker is more  likely to choose \textsc{walker}  
over \textsc{surface} message, i.e., if $p(m_1 | s) > p( m_2 | s)$,
then there are more opportunities for a phrasal $m_1$ message,
hence $c_1$ may grow faster than $c_2$. 
And in fact this will be shown to be true both numerically and
analytically:  in the limit, the phrasal utterance 
occurs with probability 1 with message $m_1$ and probability 0
with message $m_2$ when
$p(m_1 | s) > p( m_2 | s)$. Thus the speaker (in the limit) follows
the syntax that emerges from message preferences: 
she  consistently selects a  two word phrasal utterance 
for the more likely message, 
and not for the less likely message.\footnote{This section focuses on the emergence of  phrasal syntax, and not on finding a means of expression for every message.  For messages that converge to 0 speakers typically find an alternative form of expression, within the model introduced below in Section \ref{multiple-sec}.}  

\begin{thrm} 
{\bf (Fundamental Theorem:  Emergence of Semantic Composition)}

Suppose $p(m_1 | s) > p (m_2 | s)$ in the 
production algorithm for a walking event described  above.
Then, for any values of $\alpha \geq 0$, $c_1^0,\, c_2^0 >0$, 
as the number of utterances in the language history 
grows we have

a) the count ratio $c_2/c_1$ converges to 0,

b) the probability a speaker chooses a phrasal utterance 
for message $m_1$ converges to 1,

c) the probability a speaker chooses a phrasal utterance 
for message $m_2$ converges to 0.
 
\end{thrm}

\noindent
The proof of the Fundamental Theorem in
a more general form with multiple messages 
and for both speakers and hearers 
is provided in Appendix A.  

Different values for $\alpha \geq 0$ and $c_i^0 > 0$ 
will change the dynamics 
of the evolution but not the outcome. For example, 
choosing $\alpha$ large relative to the $c_i^0$ will make
phrasal utterances initially very rare, as is plausible with 
a new form.  If all $c_i^0$ are equal, then all messages
are initially equally likely to be expressed with phrases. 
One can introduce an initial bias by setting starting values $c_i^0$
unequal, but again this will not change the eventual outcome.  
We offer the robustness of this result as a model for the inevitability and ubiquity of syntax emergence.

In the Cat Walking in Grass model
the speaker chooses between exactly two potential
semantic composition rules,  where the subject (cat or grass) plays the role of \textsc{walker} or
\textsc{surface}, respectively.  
In the general model we have $I$ messages given by
$m_1, m_2, \ldots, m_I$.   
\begin{exe}\ex
\label{rich}
$p(u^{\star}_i| m_i)
= 
\displaystyle \frac{ c_i }{ c_1 + c_2 + \ldots +c_I + \alpha } $
\ \ \ \ \
\ \ \ \ 
\ \ \ \ 
\mbox{Harley-Roth-Erev formula}
\end{exe}
Theorem 1 generalizes 
as follows: if $p( m_1 | s) > p( m_i | s)$
for all $i = 2, \ldots, I$ then $c_i/c_1$ converges to 0 
and in the limit, the speaker uses a phrasal utterance 
for message $m_1$ with probability 1, and probability 0
for other messages.     
The upshot is that the syntactic construction will come to convey the most likely meaning, given the message bias.  This result obtains regardless of whether speakers choose between two or more than two possible meanings.

\subsection{Semantic interpretation by hearers}  

The language that evolves should be an effective instrument of communication.  So we now consider what a hearer understands
a phrasal utterance such as [{\it Cat walk.}] to mean, assuming they have no prior knowledge of the scene.  
They reason rationally using as input data their experience of past utterances, but they also benefit from their implicit knowledge of the message probabilities.  
Bayes's Rule can be used to estimate the probability 
that the utterance 
[{\it Cat walk.}] ($u^\star$), expresses the message 
$m_{\textsc{walker}}$ ($= m_1$).   This is shown in Appendix A.2, `The Fundamental Theorem for Hearers'.

Using the theorem  from the previous section
and assuming as before that $p(m_1|s) > p(m_2|s)$, 
we saw that $c_2/c_1$ converges to zero.  
It also follows that $p( m_1, s | u^\star )$ converges 
to 1 and $p( m_2, s | u^\star )$ converges  to 0, as shown in Appendix A.2.   
 Hence the hearer learns the syntax 
and semantic composition rule, and knows that 
{\it Cat walk.} means that the cat is the walker.  

Thus we have the second aspect of emergence
of semantics.

\begin{thrm} \label{sem-th} 
{\bf (Emergence of Semantic Interpretation) } 

Suppose $p(m_1 | s ) > p (m_2 | s )$ in the 
production algorithm for a walking event described  above.
Then, for any values of $\alpha \geq 0$, $c_1^0,\, c_2^0 >0$, 
as the number of utterances in the language history 
grows we have

a) the count ratio $c_2/c_1$ converges to 0,

b) the probability a hearer interprets a phrasal utterance 
as message $m_1$ converges to 1,

c) the probability a hearer interprets a phrasal utterance 
as message $m_2$ converges to 0.

\end{thrm}

\noindent
The generalization of this theorem to more than two possible messages 
appears in Appendix A.2.  

The speaker and hearer share the same message probabilities in the Fundamental Model.  However, in the General Model (Section \ref{variation-sec}) the message probabilities of the speaker and hearer can differ, and Theorem \ref{sem-th} (Emergence of Semantic Interpretation) still holds, as long as the hearer's probability $p(m_1)$ is greater than zero.  

\subsection{Forgetfulness improves the model}
\label{forget-sec}
Phrasal utterances, as counted in the Fundamental Model above,
are never forgotten and their value never diminished.  
But in reality people have not witnessed speech throughout the history of their language, but rather only  within their lifetimes.  
Also utterances heard recently have a greater effect than those heard
long ago \citep{Ebbinghaus:1885,murre2015replication}.  

To model  lifespan and memory limitations, we gave  
 less influence on learning to more distant memories of utterances.  
As it turns out, this modification improves the model in several important ways.  Here we explain how we implemented it. The key results are shown in the next section (Section \ref{numerical-sec}) with the help of our numerical simulations.

One can allow for the effects of forgetting by down-weighting past history with exponentially 
declining weights.  This was achieved simply by multiplying
the counts $c$ (and $\alpha$; see fn. \ref{forgetalphafn}) by a forgetting factor
$\nu$ (\textit{nu}), $0 < \nu \leq 1$,
when updating at each new utterance $u^k$. The update scheme (\ref{step3}) in Step 3 of the production algorithm in Section \ref{prod-sec} is replaced with the following: 

\begin{exe}\ex \label{forgeteq}  Updating in the Model with Forgetting for utterance $u_i^*$ (with message $m_i$):
\medskip \\
$c^k_i  = \nu c^{k-1}_i + 1  
\medskip \\
c^k_j  = \nu c^{k-1}_j , $ where $ j\neq i $
\end{exe}
Note that the payoff from phrasal utterances occurring $\ell$ phrasal utterances in the past is reduced by the factor $\nu^\ell$ while current payoff is maintained at 1.
The value of $\nu$ represents the strength of forgetting
and the smaller $\nu$ is, the faster forgetting occurs. 
Our models have $\nu$ very close to 1, with values such as 0.99 or 0.999, so that the relative impact of successive utterances differs only very 
slightly.\footnote{\label{forgetalphafn} The forgetting factor also applies to the 
$\alpha$ parameter (cp. (\ref{forgeteq})): $\alpha^k = \nu\alpha^{k-1}$.  In our numerical simulations of the Model with Forgetting (Sec. \ref{numerical-sec}), we did not apply the forgetting factor to failed utterances ($u^\dagger$).  There are many variants of forgetting models that we intend to explore in later work.} 
When $\nu = 1$ we have no forgetting and we 
recover the Fundamental Model.

The most important consequence of the model with forgetting is that it speeds up convergence.  This is intuitively plausible since recent input data is of higher quality in the sense that it is sampled closer to the convergence target. 
It would be difficult to learn contemporary English from an input randomly sampled from language stretching back to Proto-Indo-European.   
Second, it secures convergence in the unlikely event that the competing messages have equal probability.    Third, forgetting makes the outcomes stochastic (non-deterministic) in reasonable ways.  Without forgetting, convergence is deterministic,  
but with it, some languages will develop more unusual grammars, 
and stronger forgetting implies a greater likelihood of unusual convergence.  
These consequences of forgetting are discussed in the next section.

\subsection{The speed and ubiquity of convergence}
\label{numerical-sec}
The convergence theorems 
are perhaps the most important thing to know about the  
models.  But the speed of convergence and 
the factors affecting it are also important, 
and so 
we used numerical simulations to investigate them.  
Simulations can also help us get an intuitive grasp of the theory and understand why it works.

 We computed  
10,000 language histories out to 1,000,000 utterances
with three possible messages and a fixed set of 
parameters including message probabilities, 
start values ($c_i^0$),  the $\alpha$ parameter, and the forgetting factor $\nu$. 
Overall we found that the biggest factor is the message probabilities: when the two highest probabilities are far apart then the language converges rapidly on the higher of the two.  When they are close then convergence is slower, but a forgetting factor speeds up convergence in those cases.  In fact, even if the two highest probabilities are exactly the same, a forgetting factor will secure convergence.   The remaining parameters, the start values ($c_i^0$) and the $\alpha$ parameter, can delay convergence but they cannot stop it.  Overall these are very robust models of grammar emergence.

\begin{figure}[!htb] 
\vspace{-.12in}

\begin{center}
\includegraphics[width=.9\textwidth] 
{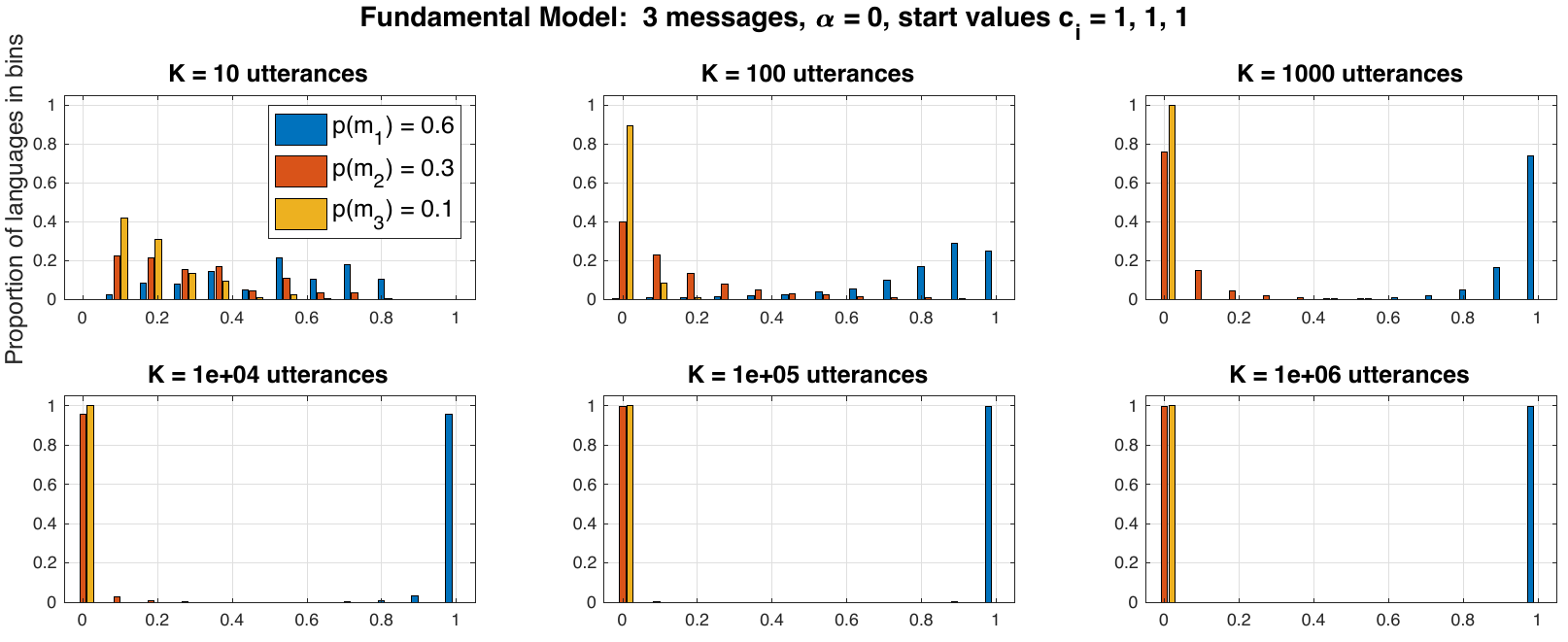}
\end{center}
\vspace{-.35in}
\begin{center}
\includegraphics[width=1.0\textwidth] 
{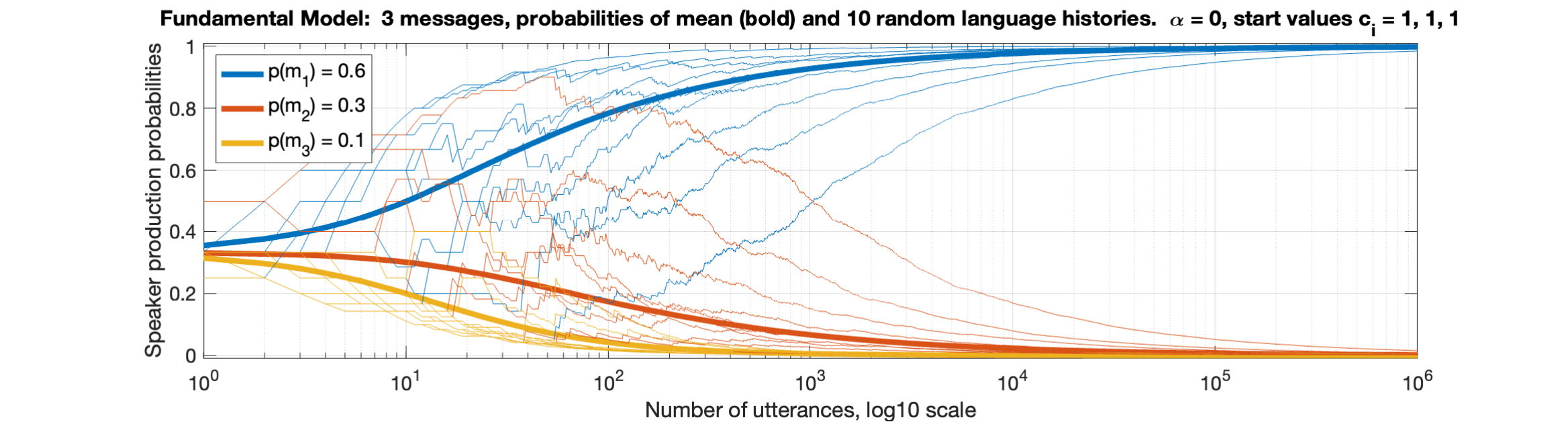}
\end{center}

\caption{Fast convergence when the probability of the first
message (blue) is larger than others.    }
\label{fig:fundamental-zero-alpha}
\end{figure}

For each simulated language at each utterance $k$
there are three phrasal sign probabilities $p(u^* | m_i)$.
If message $m_1$ converges to a point mass at 1,
then, for large $k$,  
most of the 10,000 simulated language probabilities
$p(u^* | m_1)$ will be close to
1 with other language history
phrasal sign probabilities close to 0.
 In Figure~\ref{fig:fundamental-zero-alpha}, we 
  plot probabilities of phrasal signs for the three
  messages, with probabilities 60\%, 30\%, and 10\%,  as more and more utterances are observed. This plot is on a
  $\log_{10}$ scale in order to compress the graph so we can view a
  longer timescale.  The thin lines show 10  histories selected at random from a total of 10,000 histories.
  The bold lines are mean probabilities over all
  language histories. 
  
  The bold line represents an `average language history', in that sense.  But it is important to keep in mind that all 10,000
  language histories exhibited convergence to a probability of one for the favored message.  We have included the histograms in Figure \ref{fig:fundamental-zero-alpha}, from the same simulation as the plots, to make this point dramatically.  Here probabilities are shown on the $x$-axis, separated into 12 bins.  In the last two histograms, all 10,000 languages have message $m_1$ in the highest  bin, hence very close to a probability of 1.  This result is consistent with the main conclusion of the  
Fundamental Theorem.

\begin{figure}[!htb]
\vspace{-.1in}

\begin{center}
\includegraphics[width=1.0\textwidth] 
{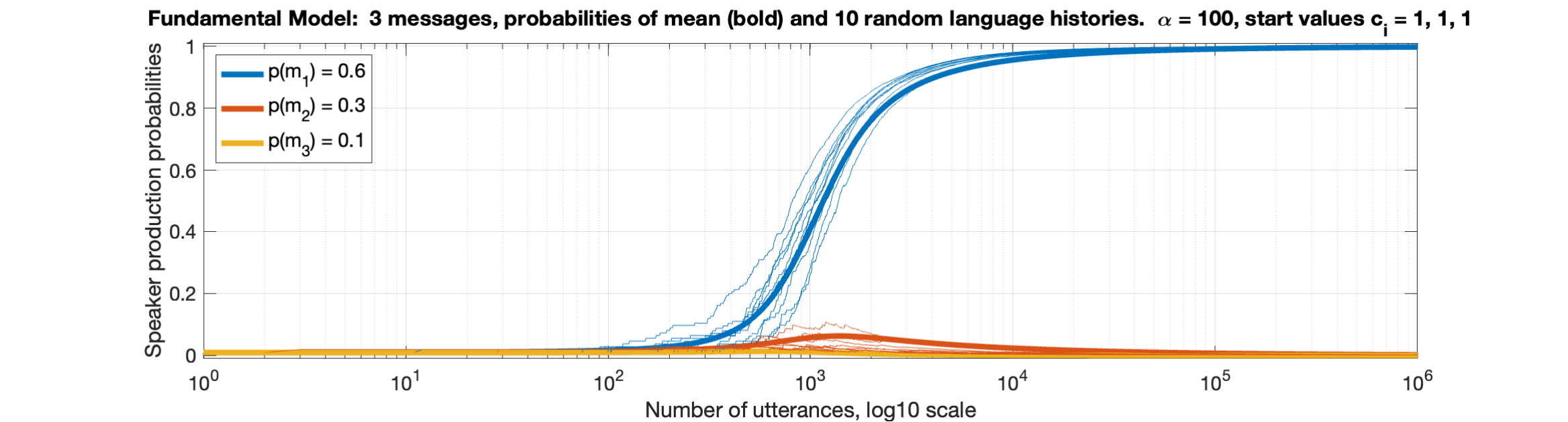}
\end{center}

\vspace{-.35in}

\caption{Initially ineffective conditioning due to a large $\alpha$
parameter.}

\label{fig:fundamental-large-alpha}.

\end{figure}

Figure \ref{fig:fundamental-large-alpha} shows the plot for the same three message probabilities, but with a large $\alpha$ parameter ($\alpha$ = 100). 
Recall that $\alpha$ is added to the denominator in the HRE formula, and so a high value depresses the overall likelihood of phrasal syntax emerging.  Comparing the plots in Figures \ref{fig:fundamental-zero-alpha} and \ref{fig:fundamental-large-alpha}, we can see that a large $\alpha$ delays the emergence of syntax--- but crucially, it does not prevent convergence.  We offer this as a model of the inevitability of the emergence of syntax.

\begin{figure}[!htb]

\begin{center}
\includegraphics[width=1.0\textwidth] 
{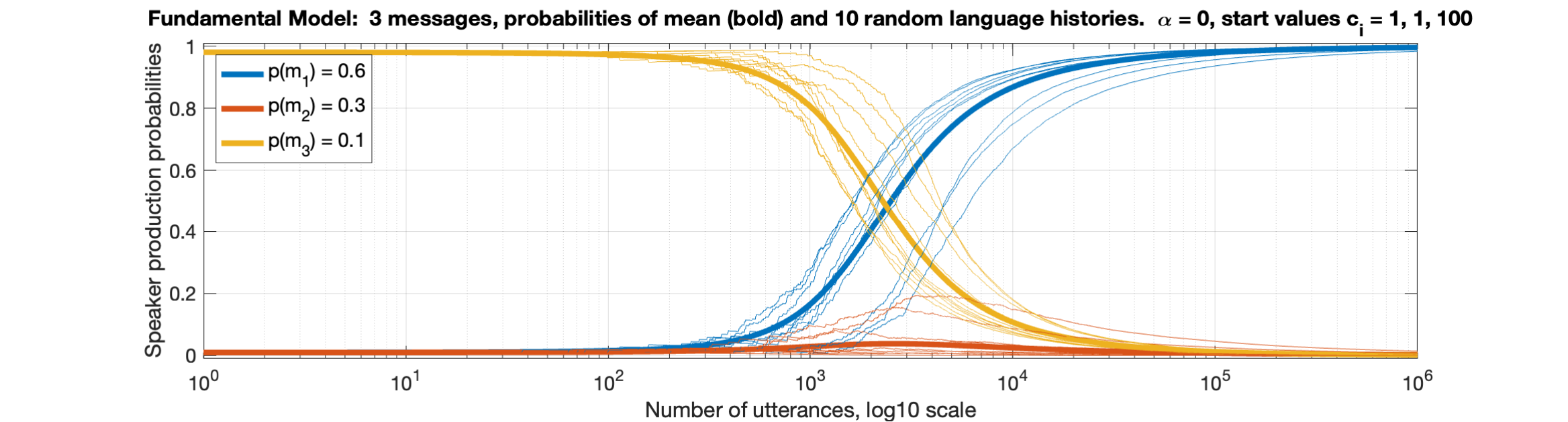}
\end{center}

\vspace{-.2in}

\caption{Large start value on message with lowest probability delays
but doesn't stop convergence of highest probability message.}

\label{fig:fundamental-large-start-value-on-lowest-msg-prob}

\end{figure}

In Figure \ref{fig:fundamental-large-start-value-on-lowest-msg-prob} we see the effect of a large start value for the `wrong' message, that is, one with a lower message probability.  This models a situation in which some  contingency leads to a temporary  interest in a normally ignored event participant, such as the grass in the Cat Walking in Grass model.  Again, this noisy start merely delays but does not stop convergence of the highest probability message.

\begin{figure}[!htb]

\begin{center}
\includegraphics[width=1.0\textwidth] 
{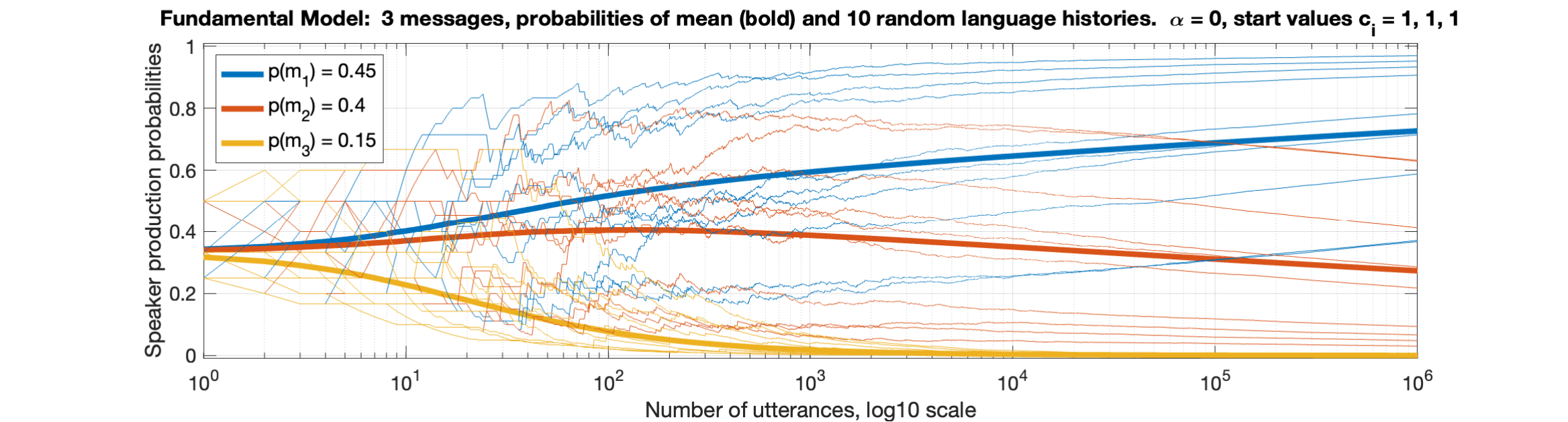}
\end{center}

\vspace{-.2in}

\caption{Very slow convergence when highest
probability messages are close.  }

\label{fig:fundamental-convergence-rate-very-slow}

\end{figure}

\begin{figure}[!htb] 

\vspace{-.1in}

\begin{center}
\includegraphics[width=1.0\textwidth] 
{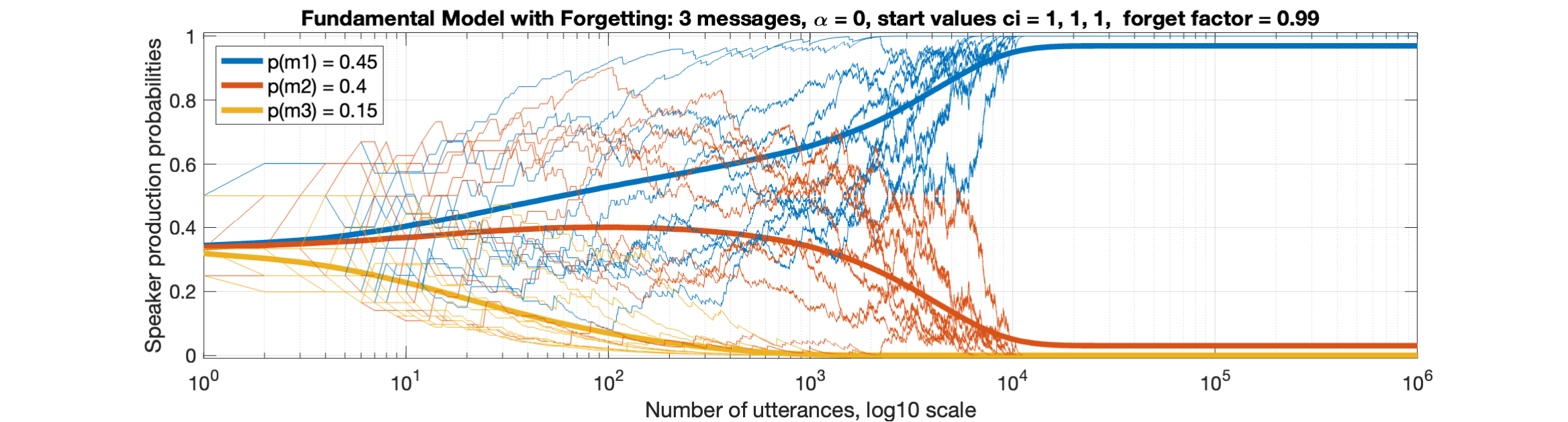}
\end{center}

\vspace{-.2in}

\caption{When the two highest probability messages are close, we get faster convergence with stronger forgetting. A few language histories
converge to the second highest probability message.  }

\label{fig:fundamental-forget-close-highest-msg-probs-990}

\end{figure}

When the two highest probabilities  are close, convergence is slow.  Figure
\ref{fig:fundamental-convergence-rate-very-slow} gives the results of a
numerical simulation with 
 message 
probabilities at 45\%, 40\%, and 15\%, hence a difference of only 5\%.  Contrast this figure with the earlier Figure  \ref{fig:fundamental-zero-alpha}, where the difference is 30\%.  
However, slow convergence can be avoided with sufficiently strong forgetting, 
as seen by comparing Figure   
\ref{fig:fundamental-forget-close-highest-msg-probs-990},
which shows the results with the same close probabilities 
(45\% and 40\%), but now with a forgetting factor of $\nu = .99$.  
This is an important result because the difference between 
the two highest probability messages is the only significant factor
affecting the speed of convergence in the Fundamental Model, and
convergence is otherwise very slow when those probabilities are close.
With forgetting, it is faster.  

When the two highest probabilities are exactly the same, then in the Fundamental Model without forgetting there is no convergence.\footnote{Technically they settle into a Beta(1,1) distribution when both initial states are equal to 1.}  
However, with forgetting, the result is very different: all languages converge on a message.  When initial states are the same then it is reasonable to expect that half of them converge to one message and half to the other, and numerical simulations support this.   The speed of convergence depends on the strength of the forgetting factor, that is, the value of $\nu$.

The theorem below shows that
all language histories 
converge, even in the case of equal highest 
probability messages.

\begin{thrm} {\bf Convergence in the Model with Forgetting}

Every language history in the the Model with Forgetting
converges to one of the messages 
$m_1, \ldots, m_I$.  

\end{thrm}
Details and proof to appear in future work.

Another interesting consequence of forgetting with close probabilities is that a few of the languages converge to the second highest probability message (see the red line in Figure \ref{fig:fundamental-forget-close-highest-msg-probs-990}).  This appears to come about when there is enough random fluctuation that in a few language histories the utterances within the memory window happen to favor the (otherwise) second highest message.  With stronger forgetting we found that more languages converge to the second highest probability message, and with a high start value for that message, we found that even more languages converge to it.  A strong forgetting factor for a particular locution might model the familiar scenario in which younger speakers intentionally `forget' the speech of their parents' generation in order to establish independence and strengthen in-group bonds.

Summarizing our findings from the numerical simulations of the fundamental model:  the main factors affecting convergence are the message probability and the forgetting factor.  Languages converge rapidly on the highest probability message, and they do so faster, the higher the probability mass on that message.   Forgetting speeds up convergence, especially when probabilities of competing messages are close; and it delivers convergence when probabilities are equal.  
In addition, forgetting can lead to convergence to a message 
that is not the most probable one.  A large initial state 
and stronger forgetting make such an outcome more likely.
Meanwhile, a lower overall likelihood of a phrasal utterance (high $\alpha$) or unusual initial conditions (high start value for the `wrong' message) delay convergence, but they still do not stop it.  

\subsection{Speaker diversity models}
\label{variation-sec}
For the Fundamental Model above we assumed that all members of the language
community have the same degree of influence and the same message biases (message probabilities).  
However, in reality some individuals are more talkative or more influential as linguistic trend-setters.  Also the impact of an utterance depends on the scene that it describes:  a warning of a bear on the attack may have greater impact on learning than a description of a harmless cat walking by.  
People also have varying interests, so the message biases could vary across speakers.  It is fairly easy to accommodate more realistic 
variability among speakers and scenes 
and still prove convergence of the learning rule.  

In an extension of the Fundamental Model that we call the \textsc{General Fundamental  Model} (or \textsc{General Model} for short), 
different members of the language community can speak more or less often with greater or lesser impact.    
We also allow scenes to influence the message 
probabilities and for scenes to have a greater or lesser impact
on learning.  
Perhaps the most striking change is that we now allow different speakers to have diverse message probability distributions, even to the point of reversing the relative probabilities of alternative messages for a given phrasal utterance.  
Despite this eclectic mix, the language community converges on the same message, within the General  Model.

The production algorithm for the General Model is almost the same as the one above
for the Fundamental Model (see Section \ref{prod-sec}), except that in
the General Model: (i) in Step 1, both a  scene ($s$) and a speaker
($r$) are chosen at random; (ii) also in Step 1, the message
probability distribution is conditioned on both the scene and the speaker
(compare (\ref{walkprob1})): 

\begin{exe} \ex \label{walkprob-r}
$1
= 
p(m_{\textsc{walker}} | s_{\textsc{walk}}, r ) +  
p(m_{\textsc{surface}}  | s_{\textsc{walk}}, r  )  $
\end{exe}  
(iii) in the final step, where the utterance counts are updated to
reflect the new utterance just produced, if $u_i$ is uttered, then
instead of adding 1 to $c_i$, we add an amount 
that varies with the speaker ($r$) and scene ($s$), in a deterministic function $\pi_i(s,r)$.  
The value of $\pi_i(s,r)$ reflects the impact  of the scene and the speaker on learning.  Hence we modify Step 3 of the algorithm for the Fundamental Model above. 
There it says that if `the utterance in Step 2 is a phrase $u^\star_i$', then  you should `add 1 to $c_i$ and then return to Step 1'.  In the General   Model we add $\pi_i(s,r)$ (instead of 1) to $c_i$ and then return to Step 1.  

The HRE formula used in Step 2 remains the same in the General Model: 

\begin{align*}
p(u^* | m_i )
&=
\frac{ c_i }{ c_1 + \ldots + c_I + \alpha}
&
\mbox{Harley-Roth-Erev rule}
\label{Harley-Roth-Erev}
\end{align*} 
Because of the change to the updating step, the values of $c_i$ are effectively weighted to reflect the  influence of different utterances varying by speaker, scene and message.  
As above, this equation gives the probability of a phrasal utterance
(represented by $u^*$) for each message $m_i$.  

The most striking feature of the General Model is that message probabilities can vary across speakers, even to the extent that speakers favor different messages the most, and yet we still get convergence on a unique message.  
Recall that 
  the Fundamental Theorem proves convergence to the highest probability message.  
For the General Model we can show that a phrase converges to the
message with the highest \textit{expected payoff}, a kind of average
over speakers and scenes.   The details of calculating the expected
payoff, as well as a theorem and proof for the General Model, will appear in future work.  The important point for grammar emergence is that the language learning model leads to conventionalization even in a diverse population.  This makes the model more realistic.  It is unnecessary to posit `an ideal speaker-hearer, in a completely homogeneous speech community' \citep[3]{Chomsky65} if we can instead model an \textit{average} speaker-hearer in a heterogenous speech community.

\section{Extensions of the Fundamental Model}

\subsection{Verbs with multiple dependents}
\subsubsection{The emergence of grammatical relations}
\label{multiple-sec}

The Cat Walking in Grass model above accounts for the emergence of a phrasal sign with a definite meaning, namely that a cat is walking, and a form of either \textit{Cat walk.} or \textit{Walk cat.}  We now posit that a minimal amount of concomitant grammatical structure emerges in order to represent that form-meaning correspondence in a way that relates it to the form-meaning correspondences of \textit{cat} and \textit{walk} alone.  That structure consists simply of a  relation between the two words. We call this particular relation subject, or $\textsc{subj}(cat, walk)$.  
An utterance of a   $\textsc{subj}(cat, walk)$ phrasal sign has the following semantic interpretation, where $\mathcal{I}$ is the semantic interpretation function:
\begin{exe} \ex \label{subjdef1}
$\mathcal{I}(\textsc{subj}(cat, walk))^k$: there is an event in $s^k$; \textit{walk}^k refers to that event, and \textit{cat}^k refers to the individual in scene $s^k$ that plays the \textsc{walker} role in that event.
\end{exe}
As for the phonological form of the phrasal sign, 
we assume for now that the earliest phrases are pronounced either [\textit{Cat walk.}] or [\textit{Walk cat.}].  Let $\mathcal{P}$  be the phonological interpretation function, let $\alpha$ and $\beta$ be the phonological strings for the words in the first and second positions, respectively, of the \textsc{subj} relation, and let $\alpha\beta$ and $\beta\alpha$ be the concatenations of those phonological strings in two orders.  Then we can state the rule for possible early forms of this phrase as the following mapping:  

\begin{exe}  \ex \label{subjdef2}
$\mathcal{P}$: (\textsc{subj}($\alpha, \beta)) \rightarrow \{ \alpha\beta, \beta\alpha   \}$
\end{exe}
$\mathcal{P}$ changes over time, a diachronic process modeled below.  Later \textsc{subj} phrases may also include case or agreement markers, auxiliary verbs and other products of grammaticalization, a process modeled in Section \ref{marked-sec}.  Also the word order often becomes fixed, narrowing the range of $\mathcal{P}$ to one order:

\begin{exe}  \ex \label{subjphon}
$\mathcal{P}$: (\textsc{subj}($\alpha, \beta)) \rightarrow \{ \alpha\beta  \}$
\end{exe}
\textsc{subj} phrases will be shown in subject-verb order (\textit{Cat walk}.) for ease of presentation, until we address the emergence of word order constraints.  

In the next subsections of the paper we systematically expand the language by adding more phrasal signs to \textsc{subj}($cat, walk$).  We do this by adding more nouns that can replace \textit{cat} (Sec. \ref{noun-sec}), more grammatical relations that can replace \textsc{subj} (Sec. \ref{gf-sec}), and more verbs that can replace \textit{walk} (Sec. \ref{similarity-sec}).  Then we introduce recursion, which allows for phrases with more than two words (Sec. \ref{recursion-sec}).  Only then will we turn to the phonological forms of signs (Sec. \ref{form-sec} and \ref{marked-sec}).

\subsubsection{Adding nouns}
  \label{noun-sec}

Let us add more nouns to replace \textit{cat}.  Note that in the Cat Walking in Grass model there need not be the same cat in every scene.  So the word \textit{cat} has variable reference, and as a variable \textit{cat} carries a restriction to cats.   With the introduction of more words for cats, it is a small step to generalize from (\ref{subjdef1}) to a rule allowing for different words for the cats (\textit{Feline walk.}) or proper names for cats (\textit{Felix walk.}).  

\begin{exe} \ex \label{Nwalk}
$\mathcal{I}$(\textsc{subj}($\langle N, walk \rangle$)): there is an event in $s^k$; \textit{walk} refers to that event, and \textit{N} refers to the individual in scene $s^k$ that plays the \textsc{walker} role in that event.
\end{exe}

Then $\langle feline, walk \rangle$ and 
$\langle Felix, walk \rangle$ are added to the \textsc{subj} set. In the Fundamental Model utterances with all those nouns (\textit{Cat walk., Feline walk.,  Felix walk}.) express message type $m_1$ ($= m_{\textsc{walker}}$ and contribute to the same count $c_1$ .  Speakers may use (\ref{Nwalk}) to generalize $m_1$:

\begin{exe} \ex \label{walker}  Some phrases expressing the grammatical relation 
\textsc{subj}($\langle N, walk \rangle$):\\
Cat walk., Girl walk., Spider walk., Centipede walk., \ldots
\end{exe}
Although they vary in their characteristic gaits, number of legs, and so on, these various walkers may be expressed with a single more general statement of the emergent \textsc{subj} relation with the English verb \textit{walk}.  
The issue of which other creatures' activities fall under the predicate \textit{walk} is a matter for word meaning that we do not address directly in this paper.  

Given (\ref{subjphon}), the word order for these new phrases is the same subject-verb order as the earlier \textit{Cat walk} phrases.

\subsubsection{Adding more grammatical relations}
\label{gf-sec}
Verbs often allow for multiple roles to be expressed.  We express the \textsc{drinker} role in \textit{Cats drink} and the \textsc{drinkee} role in \textit{Drink milk!}.  The Sequential Model introduced next accounts for multiple roles.\footnote{With the benefit of recursion (Section \ref{recursion-sec}) we can express both roles in a single sentence, as in \textit{Cats drink milk}. }

The emergence of multiple roles can be modeled with a sequence of distinct grammatical relations (GRs), each bearing an index $g = 1, 2, \ldots$ indicating its selection priority rank: the speaker first tries GR_1  for expressing their message, consulting an HRE formula; if they decide against using GR_1, they try again with GR_2, using a distinct HRE formula; and so on.   This process repeats until they either settle on a GR or run out of them.  (If they run out then they can try expressing their message another way, e.g. by adding the preposition in  \textit{Drink from bowl}.  However, we will not develop this part of the theory here.)
We illustrate with a system of two grammatical relations, GR_1 ($\equiv$ \textsc{subj}) and GR_2  ($\equiv$ \textsc{obj}).  
 
Consider the following \textit{Cat Drinking Milk} model:  speakers observe scenes of a cat drinking milk.  They describe this scene with a two word phrasal utterance that includes  \textit{drink} and either \textit{cat} or \textit{milk}.  In the beginning the language consists of utterances of the four different phrases obtained by crossing the two messages with the two grammatical relations in the sequence, \textsc{subj} and \textsc{obj} (GR_1 and GR_2).  The table below provides examples in SVO word order:

\begin{tabular}[t]{|l|l|l|l|} 
\hline
GR_g & GR name  & role & SVO exs.  \\
\hline
 GR_1  & \textsc{subj}   &   \textit{drinker}    &  Cat drink.  \\
GR_2   &  \textsc{obj}  &    \textit{drinker}   &    Drink cat.   \\
\hline
 GR_1  & \textsc{subj}   &   \textit{drinkee}    &  Milk drink.  \\
GR_2   &  \textsc{obj}  &    \textit{drinkee}   &   Drink milk.\\   
\hline          \end{tabular} 

\bigskip
\noindent
The Sequential Model production algorithm below produces languages that settle on two sentences:  the more probable of the two messages is expressed with GR_1 (\textsc{subj}) and the less probable one is expressed with GR_2 (\textsc{obj}).  If the drinker message (`a cat is drinking') is more probable than the drinkee message (`Something is drinking milk'), then the Cat Drinking Milk model settles on two phrases, \textsc{subj}-\textit{drinker} (\textit{Cat drink.}) and \textsc{obj}-\textit{drinkee} (\textit{Drink milk.}), while the other two fall out of the language.  

A speaker with the message `A cat is drinking' considers first GR_1 (\textsc{subj}), by counting up previous utterances and applying the HRE formula (\ref{subjhre}):  

\begin{exe}\ex\label{subjhre}
$p(u^\star_{\textsc{subj}} | m_{drinker}) 
= 
\displaystyle \frac{ c_{\textsc{subj},drinker} }{ c_{\textsc{subj},drinker} + c_{\textsc{subj},drinkee} + \alpha_{\textsc{subj}} } $
\hfill
 HRE formula for \textit{Cat drink.}
\end{exe}
She might utter \textit{Cat drink}.  If she does not utter it, then she considers using the object GR  and thus uttering \textit{Drink cat} instead, by applying the HRE formula (\ref{objhre}):  

\begin{exe}\ex\label{objhre}
$p(u^\star_{\textsc{obj}} | m_{drinker}) 
= 
\displaystyle \frac{ c_{\textsc{obj},drinker} }{ c_{\textsc{obj},drinker} + c_{\textsc{obj},drinkee} + \alpha_{\textsc{obj}} } $
\hfill HRE formula for \textit{Drink cat.}
\end{exe}
She might utter \textit{Drink cat}.  If she decides against it, and there are no more GRs in the sequence, then she may try expressing her message differently, such as by adding a preposition or other event modifier (see Section \ref{recursion-sec}).   

A speaker with the message `Something is drinking milk' follows the same procedure but with the following HRE formulae:

\begin{exe}\ex\label{subjhre2}
$p(u^\star_{\textsc{subj}} | m_{drinkee}) 
= 
\displaystyle \frac{ c_{\textsc{subj},drinkee} }{ c_{\textsc{subj},drinker} + c_{\textsc{subj},drinkee} + \alpha_{\textsc{subj}} } $
\hfill 
HRE formula for \textit{Milk drink.}
\end{exe}

\begin{exe}\ex\label{objhre2}
$p(u^\star_{\textsc{obj}} | m_{drinkee}) 
= 
\displaystyle \frac{ c_{\textsc{obj},drinkee} }{ c_{\textsc{obj},drinker} + c_{\textsc{obj},drinkee} + \alpha_{\textsc{obj}} } $
\hfill
HRE formula for \textit{Drink milk.}  
\end{exe}

This production algorithm appears in a general form in Appendix B.\footnote{It lacks forgetting or speaker variation but it is a straightforward matter to incorporate those.}  

As noted above, this production algorithm produces language histories that settle on two sentences:  the more probable of the two messages is expressed with GR_1 (\textsc{subj}) and the less probable one is expressed with GR_2 (\textsc{obj}).  The other two forms fall out of the language.  
This prediction can be understood by comparing the Fundamental Model above.   The \textsc{drinker} role, being the more likely one, takes the \textsc{subj} relation, just as the \textsc{walker} role did above.  Recall that in the Fundamental Model the attempt to express the second most probable role of \textit{walk} resulted in failure (\textit{Grass.  Walk.}).  In the new algorithm above, the second most probable role of \textit{drink} is given a second chance, and it gets expressed with GR_2 (\textsc{obj}).  
We illustrated a system with just two direct grammatical relations, but the production algorithm in Appendix B accommodates any number of them.


\subsection{A collective lexicon of verbs}
\label{similarity-sec}

\subsubsection{Introduction}

So far our we have derived the grammatical relations for a language with one verb, either \textit{walk} or \textit{drink}.  In this section we do the same for
 a language with more verbs.  For human language the learning algorithm in Section \ref{multiple-sec} is not fully adequate for that task.  To see why, let us try adding the word \textit{run}, and running scenes, to the Cat Walking in Grass model.  Under the models above, a speaker wishing to say that `a cat is running' uses the following HRE  formula to decide whether to say \textit{Cat run}:

\begin{exe}\ex \label{catrun}  HRE equation for producing 
$\textsc{subj}(N, run)$ (Sequential Model).  
\begin{align*}
p(u^{\textsc{subj}} | m_{\textsc{runner}})
&=
\frac{ c_{\textsc{runner}}  }{ 
c_{\textsc{runner}}  + c_{\textsc{run.surface}}  + \alpha } 
\end{align*}
\end{exe}
Suppose the speaker has witnessed many utterances of \textit{Cat walk}, but none so far of \textit{Cat run}.  
The speaker considering $\textsc{subj}(cat, run)$ for the \textsc{runner} role 
would not benefit from memories of the $\textsc{subj}(cat, walk)$ relation in \textit{Cat walk} utterances.  We might call this the \textsc{every-verb-for-itself} approach: syntax must emerge anew for each new verb.   

In contrast, human language learners acquiring the \textsc{subj} relation for one verb are influenced by the \textsc{subj} relations of other similar verbs.  One obvious piece of evidence is that the subject is expressed the same way for all verbs in a language, and that subject expression varies across languages.  There are at least two further pieces of evidence for this: 

First, consider first how learning takes place when we add a new verb to a fully developed language with many verbs.   The English transitive verb \textit{to google} was first coined in the 1990s.  Its argument mapping quickly assimilated to existing  verbs assigning roles similar to those of \textit{google}, such as \textit{look up, investigate}, and so on: googler $\to$ \textsc{subj}, googlee $\to$ \textsc{obj}.  So speakers said \textit{I googled the information} and not  \textit{*The information googled me}.  It settled on that argument structure too quickly to have depended exclusively on the message probabilities and propensities associated with the new verb \textit{google}.   Nonce word experiments confirm that learners quickly determine the argument mapping of a new verb without the benefit of usage data on the verb itself \citep{fisher1996structural}.  

Second, certain verbs have atypical message probabilities and yet they conform to the argument mapping of more typical verbs.  With a typical agent-patient verb people are more interested in expressing the agent than the patient; call such typical verbs agent-dominated.  But with some atypical verbs the patient is of greater interest.  
Speakers tend to use the passive voice for such patient-dominated verbs, since the  subject expresses  the patient argument, in the passive.   English \textit{arrest} and \textit{make} are used in passive voice more often than active, suggesting a greater interest in the patient-- perhaps due to a greater interest in identifying the suspect than the arresting officer, and a greater interest in identifying the products being made than  their makers.  Nonetheless the agent emerges as the subject (and the patient as the object) for these verbs, in the active voice.  

The goal of this section is to provide a modified  learning model for many verbs, including typical verbs, newly coined verbs like \textit{google} in the 1990s, and atypical verbs like \textit{arrest} and \textit{make}.  

\subsubsection{The Model with Similarity}
In the new approach, which we call the \textsc{collective lexicon} approach, speakers learning the \textsc{subj} relation for one verb can in principle benefit from the \textsc{subj} relations of other similar verbs they observed in past utterances.  However, they place the highest value on data involving the same verb, and proportionally less on data from other similar verbs, with a value dependent on the degree of similarity between \textsc{subj} roles of different verbs.  In terms of reinforcement learning this means that for learning a grammatical relation such as \textsc{subj}, the value of the propensities contributed by  utterances in the input depends upon the perceived semantic \textsc{similarity} between \textsc{subj} roles of different verbs; \textit{mutatis mutandis} for \textsc{obj} and other grammatical relations.  

We now update the above formula 
by including observations of not only earlier \textit{Cat run} utterances but also earlier subject-verb utterances with \emph{similar} semantic roles to the runner role, such as \textit{Cat walk}. 
Suppose the \textsc{walker} and \textsc{runner} roles have a similarity of 1/3.  Then, in computing whether to produce a subject-verb phrasal utterance of \textit{Cat run}, three past utterances of \textit{Cat walk} are equivalent to one past utterance of \textit{Cat run}.  We will express this similarity by a coefficient of 1/3 applied to the counts of subject-verb utterances with \textit{walk}, in the propensities for corresponding phrasal utterances with \textit{run}:

\begin{exe}\ex \label{Simwalkrun} HRE equation for learning 
$\textsc{subj}(N, run)$ (Model with Similarity with \textit{run/walk})
\begin{align*}
p(u^{\textsc{subj}} | m_{\textsc{runner}})
&=
\frac{ c_{\textsc{runner}} + {\frac{1}{3}}c_{\textsc{walker}} }{ (c_{\textsc{runner}} + c_{\textsc{run.surface}}) +  
{\frac{1}{3}}(c_{\textsc{walker}} + c_{\textsc{walk.surface}}) + \alpha } 
\end{align*}
\end{exe}
This is the first model with similarity that we shall adopt.  
Next we present the model in a general form and explore its predictions.

What does the similarity coefficient value reflect?   The speaker expressing the \textsc{runner} role compares it to \textit{the most similar role} of \textit{walk}.   The \textsc{runner} is more similar to \textsc{walker} than to \textsc{walk.surface} or any others, so the  coefficient value is a measure of the similarity between those roles.  
Let us restate (\ref{Simwalkrun}), rearranging the terms in the denominator to group `most similar roles' together:

\begin{exe}\ex \label{Simwalkrun2} HRE equation for learning 
$\textsc{subj}(N, run)$ (Model with Similarity with \textit{run/walk})
\begin{align*}
p(u^{\textsc{subj}} | m_{\textsc{runner}})
&=
\frac{ c_{\textsc{runner}} + {\frac{1}{3}}c_{\textsc{walker}} }{ (c_{\textsc{runner}} + \frac{1}{3}c_{\textsc{walker}} ) +  
(c_{\textsc{run.surface}} +{\frac{1}{3}}c_{\textsc{walk.surface}}) + \alpha } 
\end{align*}
\end{exe}

Let us generalize this formula for any role ($\theta_i$) of any verb ($v$), in a language with $n+1$ verbs.  Say we want to convey message/role $m_{\theta_1}$. In equation (\ref{simHRE}) a speaker is considering an utterance $u^{v\star}$ with verb $v\star$ and grammatical relation GR_g to express message $m_{\theta_1}$. 
For any verb $v \neq v\star$, we write $s_{v(\theta_i)}$ for the role of $v$ that is most similar 
to $\theta_i$ of $v\star$. Then the HRE rule with similarity will be as follows, where all the counts ($c$) are of the same grammatical relation GR_g as $u^{v\star}$, and  $c^{v\star}$ is the count of utterances with the verb $v\star$: 

\

\

\bigskip

\bigskip

\begin{exe}\ex \label{simHRE} Generalized HRE equation for a Model with Similarity \\ (All counts c represent the same grammatical relation, across different verbs.)
\begin{align*}
 p(u^{v\star} \mid m_{\theta_1}) &= \cfrac{C_{\theta_1}}{N + \alpha} \\
\smallskip \\
\text{ where } \ \ \ C_{\theta_i} &= c^{v\star}_{\theta_i} + \gamma^{v1} c^{v1}_{s_{v1}(\theta_i)} + \gamma^{v2} c^{v2}_{s_{v2}(\theta_i)} + \ldots + \gamma^{vn} c^{vn}_{s_{vn}(\theta_i)} \\
\smallskip \\
\text{ and } \ \ \ N &= \sum_i C_{\theta_i} 
\end{align*}
\end{exe}
The similarity coefficient is represented by $\gamma$ ($0 \leq \gamma \leq 1$), with a superscript $V = v1, v2, \ldots$ indicating which verb's role is being compared.  In an HRE formula for expressing \textsc{runner}, if $v2$ represents the \textsc{walker}, then $\gamma^{v2}$  represents the similarity between \textsc{walker} and \textsc{runner}.   

The learning rule \ref{simHRE} is designed to model the emergence of similar argument structures across verbs.  Next we report on how well it achieves this result and on whether language histories converge.  We will see that the new model has mostly promising results, but sometimes fails.  Then we present an improved model that solves the convergence problems while also simplifying the account of the learning process.

\subsubsection{Similarity results}
\label{decay-sec}

With atypical verbs like \textit{arrest}, speakers are more interested in the 
patient than the agent.  Without similarity, the older model wrongly predicts that the patient argument will emerge as the subject (in the active voice), producing a language with locutions such as \textit{*The thief arrested the policeman.} or \textit{*Some cloth made the weaver.}  So the old model without similarity gives the wrong result.  But does the new model shown in \ref{simHRE} give the right result?  We addressed that question through a series of numerical simulations.

We hypothesized that the Model with Similarity would have some capacity for bringing atypical verbs like \textit{arrest} into  thematic alignment with typical ones, so that the agent is correctly expressed as the subject, and the patient as the object.  
For example, consider three transitive agent-patient verbs with related meaning, \textit{stop, halt}, and \textit{arrest}.  Suppose that \textit{stop} and \textit{halt} have the typical pattern in which the agent is the most probable role (indicated by $m_1$) while the patient is the second most probable (indicated by $m_2$).  Meanwhile, \textit{arrest} has the reverse probability distribution: the patient is the more probable, the agent less.  Under the model without similarity, the predicted result is shown in the left table of Table \ref{harmonies}.  We hypothesized that under the Model with Similarity the result would instead be like the right hand table in Table \ref{harmonies}, where all three verbs have same mapping between roles (\textsc{agt/pat})  and grammatical relations (\textsc{subj/obj}).
To test this hypothesis we looked at two verbs $v_1$ and $v_2$, and varied three parameters: 

\begin{enumerate}
\item  The degree of similarity $\gamma$ between $v_1$ and $v_2$. 
\item  The relative frequency of utterances containing the typical $v_1$ and atypical $v_2$.  By definition the typical is more frequent than the atypical: e.g. $p(v_1) = 90\%, p(v_2) = 10\%$.   This is a measure of how dominant the typical argument structure is. 
\item  The difference between the two highest message probabilities, for each verb.  
\end{enumerate}

\begin{table}
\begin{tabular}{|l|l|l|}
\hline
&  SUBJ   &  OBJ\\
\hline
stop &  \textbf{m_1}:\textsc{agt}  & \textbf{m_2}:\textsc{pat}\\
halt &  \textbf{m_1}:\textsc{agt}  & \textbf{m_2}:\textsc{pat}\\
arrest &  \textbf{m_1}:\textsc{pat} & \textbf{m_2}:\textsc{agt}  \\
\hline
\end{tabular}
\quad 
\begin{tabular}{|l|l|l|}
\hline
&  SUBJ   &  OBJ\\
\hline
stop &  m_1:\textsc{\textbf{agt}}   & m_2:\textbf{\textsc{pat}}\\
halt &  m_1:\textsc{\textbf{agt}}   & m_2:\textbf{\textsc{pat}}\\
arrest &  m_2:\textsc{\textbf{agt}}  & m_1:\textbf{\textsc{pat}}\\
\hline
\end{tabular}
\caption{The verbs \textit{stop} and \textit{halt} are typical, while \textit{arrest} is atypical. 
Left: The mapping predicted by the Multiple Dependents Model (without similarity).
 For alignment each verb  follows its own message probabilities, $m$\textsubscript{1} and $m$\textsubscript{2}.  
Right: The mapping that is the hypothetical outcome of the Model with Similarity; alignment follows
 \textsc{agt} and \textsc{pat}, thematic role type.}  
\label{harmonies}
\end{table}

\noindent
For concreteness we use two agent-patient verbs, \textit{steal} and \textit{arrest}, and adopt the convention of using $v_1$ for the more frequent verb and $v_2$ for the less frequent one:

\begin{exe} \ex 
Two verb types; the most probable role of each verb is underlined:
\begin{xlista}
\ex $v_1$: Typical verb: agent is most probable role:    \underline{Man} steal money in house.
\ex $v_2$: Atypical verb: patient is most probable role:  Police arrest \underline{man} in house.  
\end{xlista}
\end{exe}
The first important result is that even with a low degree of similarity, an atypical verb (\textit{arrest}) assimilates to the typical ones (see \ref{simtest1}b):

\begin{exe} \ex \label{simtest1}
\begin{xlista}
\ex Moderate frequency difference  (70\%/30\%), moderate similarity ($\gamma = 0.3$). \\
result: The low frequency verb $v_2$ assimilates to the high frequency verb $v_1$: both verbs converge to the agent-subject mapping.
\ex  High frequency difference  (90\%/10\%), low similarity ($\gamma$ = 0.1).\\
result: The low frequency verb $v_2$ assimilates to the high frequency verb $v_1$: both verbs converge to the agent-subject mapping.
\end{xlista}
\end{exe}
We tested verbs with very low similarity ($\gamma$ = 0.03).  The atypical verb failed to converge to a mapping.  However, with the addition of a forgetting factor, even low similarity verbs converged, the atypical ones assimilating to the typical ones:  

\begin{exe} \ex  \label{simtest2}
\begin{xlista}
\ex \label{fail} Moderate frequency difference  (70\%/30\%), very low similarity: ($\gamma$ = 0.03). \\ 
result: The atypical verb $v_2$ converges to intermediate probabilities (failure).
\ex Same as (\ref{fail}), but add a weak forgetting factor: \\
result: The atypical verb $v_2$ assimilates to the high frequency verb $v_1$: both verbs converge to the agent-subject mapping.
\end{xlista}
\end{exe}
These initial results were promising.  However, we were unable to demonstrate convergence within a reasonable time under all conditions.  
Next we analyze the problem and propose a solution.

\subsubsection{Decaying similarity improves the model}
To understand  the conditions leading to very slow convergence, consider the different consequences of a similarity coefficient $\gamma$  close to zero, and close to one.  
If $\gamma $ is at zero, we recover the model without similarity, and so 
with a patient-dominant verb, the patient emerges as subject.  If $\gamma $ is very close to zero, then the result is the same.  
At the other extreme, if $\gamma$ is at one, then it is strongly affected by the typical agent-dominated verbs, and the agent emerges as subject.  Most values  between `close to 0' and `close to 1' have the same result, the agent emerges as subject.   But there are values of $\gamma $ on the cusp between those two states, where we get no convergence within a reasonable time.  Speakers remain in a state of perpetual indecision, unable to adjudicate between conflicting evidence.

We solved this problem by imposing a decaying factor on similarity, somewhat analogous to forgetting but now diminishing the similarity coefficient $\gamma$ over time.  
This places a statute of limitations on the pressure on atypical verbs to conform to the typical ones.  Any verbs that resist the pressure to conform for long enough are eventually left to go their own way.  The motivation behind decaying similarity is that once the argument structure of a given verb is established, learners no longer need to consult data from other verbs.  
As with forgetting, the decaying factor makes learning easier by directing the learner's attention to the most useful input data.  
It is simpler and more effective.  
In fact we found that a Model with Similarity that has both forgetting and decaying similarity \textsc{always leads to convergence in reasonable time}.

\begin{figure}   
 \begin{center}
 \includegraphics[width=.5\textwidth] {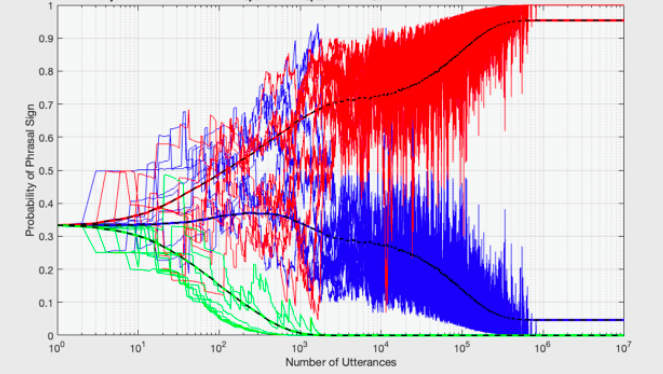}
 \includegraphics[width=.48\textwidth] {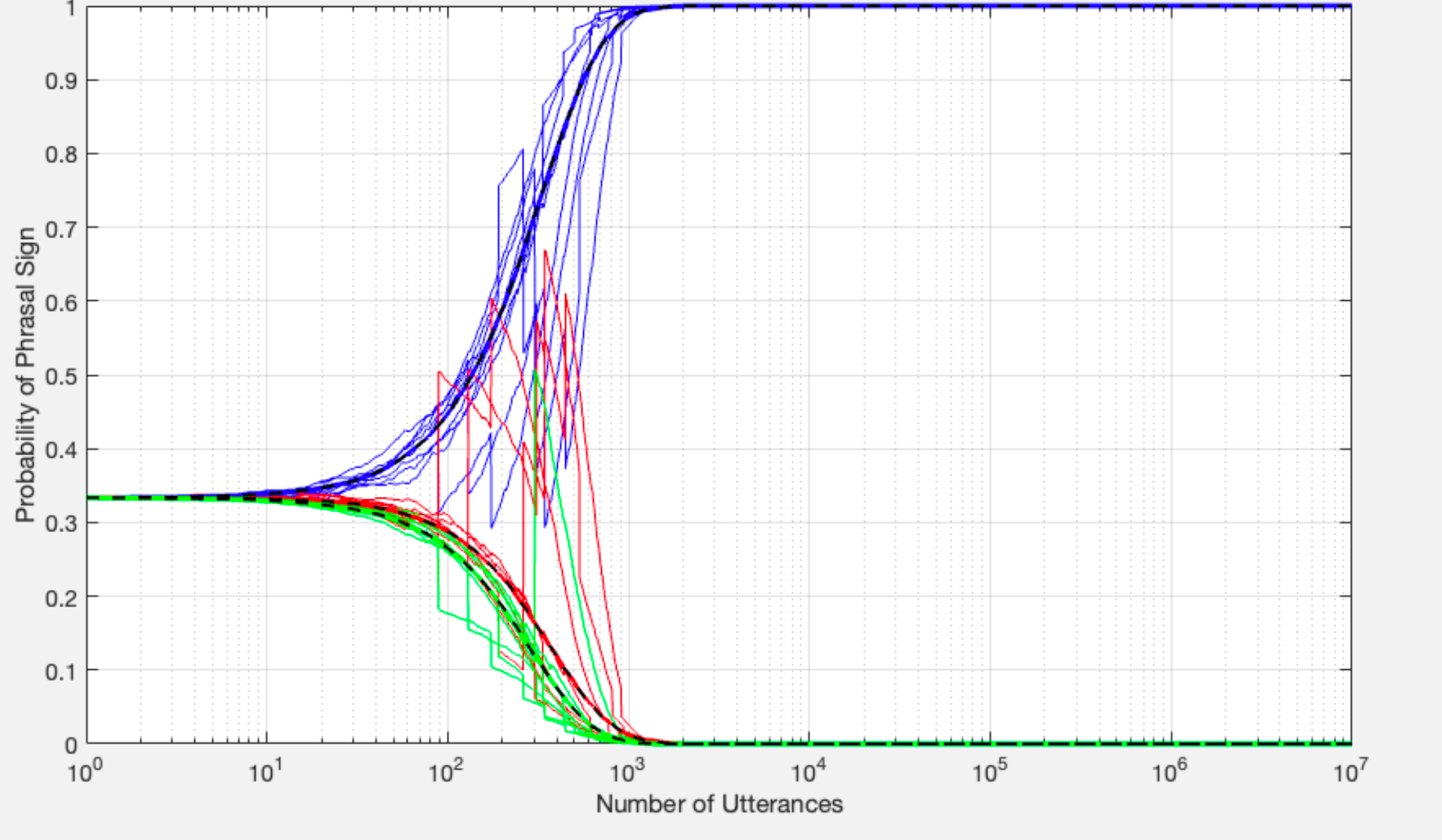}\\
 \medskip 
\begin{tabular}{|r|l|l|l|}
\hline
&  v_1   &  v_2 & $\gamma$ \\
& \small \textit{steal} & \small \textit{arrest} & \\
\hline
p(v) &  0.7 &  0.3   & \\
\hline
red: p(m_{\textsc{agt}})  & 0.7 &  0.3 & .03 \\
blue: p(m_{\textsc{pat}})  & 0.2 &  0.5  & .03\\
green: p(m_{\textsc{loc}})  & 0.1 &  0.2  & .03\\
\hline
\end{tabular} 
\quad   \   \, \, \, \
\begin{tabular}{|r|l|l|l|}
\hline
&  v_1   &  v_2 & $\gamma$ \\
& \small \textit{misc.} & \small \textit{google} & \\
\hline
p(v) &  .99  &  .01   & \\
\hline
blue: p(m_{\textsc{agt}})   & 0.7 &  0.3 & .01 \\
red: p(m_{\textsc{pat}})  & 0.2 &  0.5  & .01\\
green: p(m_{\textsc{loc}})  & 0.1 &  0.2  & .01\\
\hline
\end{tabular}
  \end{center}
\caption{Plots of 2000 language histories with averages and 8 sample plots, showing the probability that each role of verb $v$\textsubscript{2}, agent, patient, or location, is expressed as the subject of $v$\textsubscript{2}, in the Model with Forgetting and Decaying Similarity.  The table beneath each figure shows its probability matrix, with relative frequency $p(v)$ of $v$\textsubscript{1} and $v$\textsubscript{2} utterances; message probabilities for $v$\textsubscript{1} and $v$\textsubscript{2}; and role similarity coefficient $\gamma$.  All roles in all 2000 language histories converge to 1 or 0, in both simulations.  Note that in the plot on the left, a few blue roles converge to 1 and a few red roles to 0, so the averages, shown by the thick lines, converge to constants near 1 and 0.}   
\label{decay-fig}

\end{figure}

Figure \ref{decay-fig} shows the results of two studies of the Model with Forgetting and Decaying Similarity.   
Both simulations are Models with Forgetting and Decaying Similarity, producing 2000 language histories.  Each simulation had two 3-role verbs ($v_1$ and $v_2$), and all roles converge to either 1 (express this role as the subject) or 0 (do not express this role as the subject).  Both plots show the outcome for the atypical verb $v_2$, and not for typical verb $v_1$.  Message probabilities for each simulation are shown in the table below it.  

The plots on the left show the results of a simulation of an \textit{arrest}-type atypical verb with a decaying similarity factor of 0.9999 and other parameters as shown below the plot.  The plot shows the history of verb $v_2$ \textit{arrest} (see the table below the plot).  Importantly, all 3 roles in all 2000 language histories converged to 1 or 0.  The thick lines showing the averages are not quite at 1 and 0, because in a few languages the atypical verb has resisted the pressure to conform and the patient has been selected over the agent as the subject.  

The plots on the right show the results of a simulation of a \textit{google}-type verb coinage  in a language with many verbs, with a decaying similarity factor of 0.99999.   
The plot shows the history of verb $v_2$, which accounts for only 
1\% of utterances, while the other 99\% have verb V_1.  This simulates the notion of a coinage entering a large lexicon in which the vast majority of verbs (99\%) conform to the typical agent-subject pattern.  The newly coined verb, which is assumed to be patient-dominant in order to test the theory, conformed to the typical pattern within a reasonable time, in all 2000 language histories.  

We have used low similarity coefficients (.03 and .01) to show the robustness of the model.  With higher coefficients, convergence is faster.  In conclusion, the Model with Forgetting and Decaying Similarity accounts for typical verbs, newly coined verbs like \textit{google} in the 1990s, and atypical verbs like \textit{arrest} and \textit{make}.
Next we situate our Model with Similarity relative to other work in psychology and linguistics.

\subsubsection{Similarity in language and learning}

The Model with Similarity is a new mathematical model, but the principles underlying it are consistent with prevailing views on learning and language acquisition, and the model predictions are broadly consistent with cross-linguistic generalizations emerging from descriptive studies of language.  

Let us first consider the notion of similarity itself.  In frequency-based learning, the learner deems a new observation to be sufficiently similar to earlier ones retrieved from memory, to support the conclusion that the name applied to the earlier ones should apply to the new one as well.   Every cat is different, and learning whether the creature before us should be called a \textit{cat} depends on its perceived similarity to previous cat observations.  
 For that reason similarity judgments play a fundamental role in psychological theories of word meaning and concept formation \citep{murphy2004}.

There are various theories of how similarity judgments are applied.  For many concepts people distinguish between better and worse instances, and according to \textsc{prototype theory} people do this by judging the relative similarity of an instance to a central summary representation, the \textsc{prototype}  
\citep[for an overview see Murphy 2004]{rosch1975family, rosch+etal:1976,rosch1978principles}.  
  In \textsc{exemplar theories}, instead of forming a single prototype, one holds in memory all the exemplars encountered.   \citet{nosofsky1986} posits a mechanism for judging concept membership for a new object in which you compute similarity to stored exemplars of different concepts, and  the new item is judged based on its most similar neighbor among the remembered items.  
In the \textsc{knowledge approach} (also called the theory view or the theory theory), similarity judgments for categorization depend upon a richly structured knowledge base and cannot be computed in isolation from it
 \citep[ Ch. 6]{murphy2004}.

These insights into the acquisition of 
 word meaning have been extended to the acquisition 
 of syntax as well \citep[ inter alia]{Tomasello:2003,Goldberg:2006,Goldberg:2019}.   
That is a central idea behind \textsc{constructional} approaches to the acquisition of syntax: syntactic constructions carry some meaning, like words, and so a construction is chosen to express a meaning similar to the meanings of previously heard locutions involving the same construction type.   Our Model with Similarity belongs within this family of approaches.  The term \textsc{construction} is sometimes used for the clusters of similar argument structures that emerge in acquisition models related to ours, such as the \citet{alishahi+stevenson:2008} computational learning model (`A\&{S}'): 
 \begin{quote}
In the A\&{S} model, constructions are viewed simply as a collection of similar verb usages. Each verb usage, represented as a \textit{frame}, is a collection of features which can be lexical (the head word for the predicate and the arguments), syntactic (case marking, syntactic pattern of the utterance) or semantic (lexical characteristics of the event and its participants, thematic roles that the participants take on). A construction is nothing more than a cluster of such frames.  \citep[81]{Alishahi:2014}
\end{quote}
The Model with Similarity is consistent with this general type of theory of the  clustering of `similar verb usages'.  We represent the output of such similarity calculations with a single coefficient; and our scenes correspond to their frames.  However, our Model with Similarity differs from the one described in the above quote in that our notion of similarity is exclusively semantic and not syntactic.  Formal syntactic features such as `case marking, syntactic pattern of the utterance', and so on do not enter into determining similarity. Instead the formal side of a cluster is idealized to a relational invariant such as \textsc{subj} or \textsc{obj}.  Variations in form are handled by two other components of the theory, the Model with Forms and the Form Competition Model.  

The similarity coefficient $\gamma$ is non-negative.  Data from one verb can encourage another to assimilate, but one verb cannot inhibit another.  As a consequence, multiple distinct semantic role clusters can form for a given grammatical relation.   But is there evidence for such clusters in human language?

In fact semantic role similarity clusters of this kind have been a mainstay of grammatical description and theory for thousands of years.  
In the 4th Century BC 
the Sanskrit grammarian P\={a}\d{n}ini  took note of them and described them with a system of \textsc{thematic role types} called \textit{k\={a}rakas}.  \citep{kiparsky+staal:1969}.
Among others they included
\textit{ap\={a}d\={a}na} (source), 
\textit{karman}  (object of desire),  \textit{kara\d{n}a} (instrument), 
\textit{adhikara\d{n}a} (locative), \textit{kart\d{r}} (agent), and \textit{hetu} (Cause).   
This approach recurs often throughout the history of grammatical study.
Some studies consider the full ensemble of roles associated with a given set of grammatical relations of a verb in an utterance, the verb's \textsc{predicate argument structure}.  Semantic similarity clusters classified by predicate argument structure are sometimes simply called \textsc{verb classes} \citep{levin:1993}.
\citet{fillmore:1968,Fillmore:1977} observed that the selection of a role type for expression as the subject of a verb is governed by  a hierarchy of preference.  Subject preference rules take the following form: in a given predicate argument structure, if there is an agent, it becomes the subject; otherwise if there is a beneficiary, it becomes the subject; otherwise, if there is an experiencer or recipient, it becomes the subject; and so on, for the remaining role types in a ranked ordering hierarchy such as (\ref{thetahierarchy}) (this version of the hierarchy is from  \citet[329]{Bresnan+etal:2015}): 

\begin{exe} \ex \label{thetahierarchy} A thematic hierarchy of preference for subject selection \\
agent > beneficiary > experiencer/recipient > instrument > patient/theme > locative
\end{exe}
On the present models such generalizations are predicted  to emerge from the \textit{typical} message probability distribution over thematic role types.  Suppose that in a typical verb with an agent participant, the agent is most likely to be mentioned.  Then the agent is predicted to emerge as the subject, by the Fundamental Theorem.  There are also  verbs with atypical message probabilities, where a non-agent is more likely to be mentioned than the agent.  But the atypical verbs assimilate to typical verbs, in the Model with Similarity.  As a result all verbs are theoretically predicted to conform to a single thematic hierarchy of preference for subject selection.

\subsection{The Model with Recursion}
\label{recursion-sec}
Natural languages have sentences with many more than two words.  
This suggests that speakers not only combine words into phrases but also combine phrases with words and with other phrases.  It is a simple matter to adjust the model to allow this.  In Step 2 of the production algorithm above, the speaker chooses two \textit{words} to combine; in the revised Model with Recursion, the speaker may choose words or phrases, with \textit{sign} as the general term encompassing both.   For simplicity we limit the phrasal signs that may combine with other signs to the ones that have already converged.  The term \textit{known signs} will be used for the union of the set of words in the lexicon and the set of converged phrasal signs.   So in Step 2 of the production algorithm, \textit{word(s)} is replaced with \textit{known sign(s).}  
 
To illustrate, assume a lexicon of three words, \textit{cat, walk}, and \textit{grey}.  The phrase [\textit{grey cat}] emerges in a process similar to that of [\textit{cat walk}].
The word \textit{grey} describes greyness in one of two possible semantic roles: as the color of the cat ($m_f$); or  as the color of the area around the cat ($m_s$).  On meaning $m_f$ [\textit{grey cat}] describes a cat with grey fur, while on meaning $m_s$ it describes a cat (of any color) lying on a grey stone.  Meaning $m_f$ is the more popular so by the Fundamental Theorem it emerges as the interpretation of [\textit{grey cat}], it converges and becomes a known sign.   
As a known sign [\textit{grey cat}] can replace \textit{cat} in the algorithm in Section \ref{prod-sec}.  The result is a three word sentence containing the words \textit{grey, cat}, and \textit{walk}, meaning `A grey cat is walking.'  

As a second example of recursion, consider how a language could develop transitive verbs.  The phrases  [\textit{cat drink}] and [\textit{drink milk}] were derived above.  If [\textit{drink milk}] is a known sign it can replace \textit{drink} in the subject-verb rule, resulting in [\textit{cat}  [\textit{drink milk}]].  

As a third example, consider the modification of one verb by another in a serial verb construction, as in this Thai sentence:
\begin{exe}
\ex \label{thai}
\gll	Piti	den	pay	th\v{\textturnm}{\textipa{N}}	  roo{\textipa{N}}rian.\\
	Piti	walk	 {go.there} 	arrive	school	\\
	\glt	 `Piti walked to school.'
\end{exe} 
Each of the verbs \textit{den}	`walk', \textit{pay}	`go.there', and	 
\textit{th\v{\textturnm}{\textipa{N}}} `arrive' can appear on its own in an independent clause.  When serialized as in \ref{thai}
they describe a single event.  Note that the Thai verb \textit{th\v{\textturnm}{\textipa{N}}} becomes `to' in the English translation line.  This kind of verb modifier can develop in various directions, one of which is to become a preposition, and then a case marker.  This is shown schematically in (\ref{affect}).   We start from serial verbs (Stage I).  One verb becomes an adposition like \textit{to}, which marks a thematic role type of the verb (Stage II).  
Then the adposition can lose its semantic content and thereby become simply a marker of the grammatical relation, that is, a case marker.  The marker of the object relation is shown here as ACC for accusative case  (Stage III).  In a final step the accusative marker becomes an affix (Stage IV).  

\begin{exe}
\ex \label{affect} Stages in the development of a case marker \\
I. Dog bite_V affect_V cat.  \hfill  serial verbs\\
II. Dog bite_V  [ affect_P cat ]. \hfill       P retains some content, marks thematic role type \\
III. Dog bite  [ \textsc{acc}_P cat ].	\hfill	 P loses its content, marks the OBJ of \textit{bite}\\
IV. Dog bite   cat-\textsc{acc}	\hfill	 morphologization \\
\end{exe}
In Section \ref{marked-sec} we model the process by which a case marker becomes obligatory.

\subsection{The Model with Forms}
\label{form-sec} 

Having focused so far on the emergence of grammatical relations, we now finally turn to the morpho-syntactic forms of sentences, the phrasal structures and functional morphemes that express those grammatical relations.   We split this task into two parts.  First we show how a word order can come to express the subject relation within an extension of the Fundamental Model called the Model with Forms.  Then in Section \ref{marked-sec} we show how other forms for the expression of grammatical relations emerge within a new model called the Form Competition Model. 

Many human languages use word order to indicate grammatical relations in the clause.   Of them, about 90\%  have sentence-initial subjects, that is, they have either SOV or SVO order (see Table \ref{sovtable}).\footnote{Word order expression of the object relation is treated under the Form Competion Model (Section \ref{marked-sec}), so that different forms can `compete' for expression of the object.}  We illustrate the Model with Forms by accounting for the emergence of a sentence-initial subject position.  

\begin{table}\centering
 \begin{tabular}{l|l|l}
   & n & \% \\
 \hline
SOV &	564	& 47 \\
SVO &	488	& 41 \\
VSO &	95	& 8 \\
VOS &	25	& 2 \\
OVS &	11	& 0.9 \\
OSV &	4	& 0.3 \\
\end{tabular}
\caption{\label{sovtable}Sample of languages with a dominant word order. Adapted from \citet{Dryer:2013}.}
\end{table}

In the model below we posit a production bias favoring the expression of subjects early in the utterance.  The  bias derives from a well-established finding from language processing: words that are easier to retrieve from memory appear earlier in utterances.  This so-called \textsc{easy-first} generalization, which has been related to memory retrieval, motor planning, and serial order in action planning,  ``has enormous influence on language form,'' according to \citet[3]{MacDonald:2013}.  Words and phrases that are easier to access are more predictable, shorter, less syntactically complex, and \textit{more likely to be previously mentioned in the discourse} \citep{levelt1982,Bock+Warren1985,tanaka+etal:2011}.  An individual with the last of these properties, being previously mentioned, is called a \textsc{discourse topic}.  It has been independently observed that the subject referent is the most likely event participant to be the discourse topic (\citet{li1976subject}, \citet{Andrews:1985}, \citet[100]{Bresnan+etal:2015}).  Summarizing, a subject-first bias may follow from the easy-first generalization, together with a tendency for subjects to be topical and therefore easy to process.\footnote{A  related \textsc{agent-first} generalization is supported by gesture studies. When asked to describe a scene with gestures, people tend to sign the agent participant before signing the event type or other participants \citep{Goldin-Meadow-etal:2008, Gibson+etal:2013, Hall-etal:2013, Hall-etal:2015, Futrell+etal:2015}.}

Consider the language mentioned in Section \ref{recursion-sec} with the following known signs:  \textit{cat, drink, milk}, and [\textit{drink milk}].   Speakers can express the \textsc{subj} relation in either the subject-predicate order in (\ref{subjpred}.I) or the predicate-subject order shown in (\ref{subjpred}.II):

\begin{exe}
\ex\label{subjpred} Expressing the \textsc{subj} relation:\\
I. subject-predicate order ($f_{s.p}$): 
 \begin{xlista}
\ex Cat  drink.
\ex Cat [drink milk].
\end{xlista}
II. predicate-subject order ($f_{p.s}$): 
 \begin{xlista}
\ex   Drink cat.
\ex  {[Drink milk] cat.}
\end{xlista}
\end{exe}
Let us assume fixed universal probabilities for word order in 
a \textsc{subj} phrase.  Let $f_{s.p}$ represent the form with subject-predicate order,
and let $f_{p.s}$ represent the form with predicate-subject order.  Given the easy-first generalization, we may assume subject-predicate order is more frequent than predicate-subject order, so we have the following probabilities for the expression of the \textsc{subj} relation:

\begin{exe}\ex \label{subj1dist}
\begin{eqnarray*} 
1
&=&
p( f_{s.p} |  m_{\textsc{subj}}) 
+
p( f_{p.s} |  m_{\textsc{subj}}) 
\\
p( f_{s.p} | m_{\textsc{subj}}) 
&>&
p( f_{p.s} |  m_{\textsc{subj}}) 
\end{eqnarray*}
\end{exe} 
Following Step 2 of the production algorithm in Section \ref{fundy-sec}, a new step  is added: \\

\noindent
 {\bf Step 2a. Select a form.} If the speaker has chosen not to utter a phrasal \textsc{subj} sign, then skip this step.  If speaker has chosen to utter a phrasal sign, then she chooses a word order 
using the following counts.
\begin{exe}\ex Counts
\begin{eqnarray*} 
c_{\textsc{subj}}
&=&
\mbox{ utterances of phrasal signs with \textsc{subj} relation } 
\\
c^{s.p}_{\textsc{subj}}
&=&
\mbox{ utterances of subject-predicate order phrasal signs with \textsc{subj} relation }  
\\
c^{p.s}_{\textsc{subj}}
&=&
\mbox{ utterances of predicate-subject order phrasal signs with \textsc{subj} relation }  
\end{eqnarray*}
\begin{eqnarray*} 
c_{\textsc{subj}}
&=&
c^{s.p}_{\textsc{subj}}
+
c^{p.s}_{\textsc{subj}}
\end{eqnarray*}
\end{exe}
We now define 
a frequency-based learning/production formula analogous to equation \ref{rich} above.
The probability the $k$th utterance is $u^{s.p}$ 
(subject-initial)
and the probability $k$th utterance is $u^{p.s}$ 
(verb-initial) given message $m_{\textsc{subj}}$
are given by the following.
\begin{exe}\ex \label{richformS} 
\begin{align*}
p( u^{s.p} | m_{\textsc{subj}} )
 \ = \
\frac{ c^{s.p}_{\textsc{subj}} }{ c^{s.p}_{\textsc{subj}}+ c^{p.s}_{\textsc{subj}} + \alpha }
p( f_{s.p} | m_{\textsc{subj}}  )
\\
p( u^{p.s} | m_{\textsc{subj}} ) 
 \ = \
\frac{ c^{p.s}_{\textsc{subj}}}{  c^{s.p}_{\textsc{subj}}+ c^{p.s}_{\textsc{subj}}  + \alpha }
p( f_{p.s} | m_{\textsc{subj}} )
\end{align*} 
\end{exe}
The Model with Forms shares the same structure  with
the Fundamental  Model and thus has analogous convergence
properties:  we get convergence to the
 form associated with product
$p ( f_{s.p} | m_{\textsc{subj}} )p(m_{\textsc{subj}}) $ or 
$p ( f_{p.s} | m_{\textsc{subj}} )p(m_{\textsc{subj}}) $  
with highest probability. 
All other utterance forms 
converge to 0 in probability.
A more general result with more than two forms 
holds. The theorem and proof are in Appendix C. 

Upon convergence the resulting language has subject-predicate structures like (\ref{subjpred}.I) and  lacks predicate-subject structures like (\ref{subjpred}.II).


\section{The Form Competition Model}
\label{marked-sec}
\subsection{Grammaticalization}

In the models presented so far, each element of the grammar that emerges is directly motivated by the speaker's intention to express a meaning.  The Multiple Dependents Model gave us a system of grammatical relations, with \textsc{subj} and \textsc{obj}, but we still need to distinguish between their expressions.  Human languages use function morphemes and word order rules for that task, and so we will show how they emerge.  
In this section we show  how  function morphemes and word order constraints become grammatically obligatory, and how this enables them to form complex grammatical systems.
We will illustrate first with a story of how accusative case markers go from being optional to obligatory, for the expression of the \textsc{obj} relation.

\subsection{When forms compete}
\label{cat-sec}
How do accusative case markers  become obligatory for objects?   Suppose accusative case has emerged in a language without fixed word order, as described in Section \ref{recursion-sec}.  But the accusative case marker is optional, so a sentence like (\ref{bitten}a) is ambiguous, with alternative interpretations that depend on whether \textit{cat} is subject or object:

\begin{exe} 
\ex \label{bitten}
\begin{xlista}
\ex  Cat bite.  \\ 
\textsc{subj}: `The cat is biting (something).'  \\
\textsc{obj}:  `Something is biting the cat.' \\ 
\ex  Cat-\textsc{acc} bite.  \\
\textsc{obj}: `Something is biting the cat.' 
\end{xlista}
\end{exe}
In the system shown in (\ref{bitten}), the two forms \textit{cat} and \textit{cat-}\textsc{acc} are competing for expression of the \textsc{obj} message.  If the 
marked form \textit{cat-}\textsc{acc} wins out, to the exclusion of \textit{cat},
then the result is an efficient system with a one-to-one mapping between forms and grammatical relations.  

That historical scenario is shown concisely in Table \ref{sotable}.
In Stage I the unmarked form \textit{cat} is used ambiguously for either the \textsc{subj} or \textsc{obj}.  
 We follow \citet[20]{deo2015} in referring to the transition from Stage I to II as \textsc{recruitment}: a precursor to an accusative case marker is `recruited' to mark (some) objects but not subjects, in a process described in Section \ref{recursion-sec} above.  This gives us the situation in Stage II.  We refer to the transition from Stage II to III as \textsc{categoricalization}, because the case rules become obligatory or categorical.  
Note that at Stage III the unmarked form (\textit{cat}, as opposed to \textit{cat}-\textsc{acc}) can only express the subject, not the object, so in that sense the zero-marked form \textit{cat} has become a nominative case form at Stage III.  
 
\begin{table}\centering
\begin{tabular}{l|c|c}
&  unmarked   &  \textsc{acc} \\
\hline
Stage I &  $\textsc{subj} \vee \textsc{obj} $  & \\
Stage II  &  $\textsc{subj}  \vee \textsc{obj} $ &  $\textsc{obj} $ \\
Stage III  &  $\textsc{subj} $ &   $\textsc{obj} $ \\
\end{tabular}
\caption{\label{sotable}Recruitment and categoricalization.}
\end{table}

The Form Competition Model presented below is a model of the transition from Stage II to III.  As we will see, the model predictions depend upon the message probabilities: if \textsc{subj} is more likely than \textsc{obj} overall, then we predict that the accusative case marker will become obligatory for objects.  However, if \textsc{obj} is more likely than \textsc{subj} then we predict that the alternation between forms will continue, with a theoretical predicted relative frequency that depends only on the relative likelihood of the messages.

\subsection{A production model for competing forms}
 \label{fcmodel-subsec}

For our first model we assume a message bias whereby subjects of transitive verbs (\textsc{subj}) are more likely to be expressed overall than objects 
 (\textsc{obj}). 
  Letting $m_{\textsc{subj}}$ and $m_{\textsc{obj}}$ represent the two messages shown in \ref{bitten}, we have:

\begin{exe} 
\ex \label{biter} \ \ \ \ \ \ 
$p(m_{\textsc{subj}}) > p(m_{\textsc{obj}})$
\end{exe}
We are currently at Stage  II, so a speaker can express the latter message ($m_{\textsc{obj}}$) two different 
ways, namely with or without the \textsc{acc} marker.
We will use the following abbreviations (here subscripts distinguish messages, while superscripts distinguish forms.):
\begin{itemize}
\item $f^u$ (`unmarked form'):  [Cat bite]. 
\item $f^a$ (`accusative form'):  [Cat-\textsc{acc} bite]. 
\item $m_{\textsc{subj}}$: `The cat is biting.'
\item $m_{\textsc{obj}}$:    `Something is biting the cat.' 
\end{itemize}
Forms $f^u$ and $f^a$ are the only ways to express the object, hence:
\begin{exe} 
\ex \label{onlyfufa} \ \ \ \ \ \ 
$1 = p(m_{f^u|\textsc{obj}}) + p(m_{f^a|\textsc{obj}})$
\end{exe}
This model will be demonstrated with a production algorithm and a theorem.  

\

\noindent
{\bf Production algorithm for form competition}\\

{\bf Step 1} The speaker selects a message, 
$m_{\textsc{subj}}$ or $m_{\textsc{obj}}$, with fixed probabilities:

\begin{exe}[equation]
\ex 
$1 =
p(m_{\textsc{subj}}) + p(m_{\textsc{obj}}) $
\end{exe}

{\bf Step 2} The speaker chooses between producing form $f^u$ or $f^a$, using counts of previous utterances.

\

 \ \ $c^u_{\textsc{subj}}$ = count of utterances with unmarked form $f^u$ and the meaning  $m_{\textsc{subj}}$ 

 \ \ $c^u_{\textsc{obj}}$ = count of utterances with unmarked form $f^u$ and the meaning  $m_{\textsc{obj}}$

 \ \ $c^a_{\textsc{obj}}$ = count of utterances with marked form $f^a$ and the meaning  $m_{\textsc{obj}}$

\

\noindent
Taking $m_{\textsc{subj}}$  first, the bare noun $f^u$ is the only option, since the \textsc{acc}-marked form $f^a$  is specialized for the affected theme of the action, the object of the verb:  

\begin{exe}\ex  Expression of the subject with the unmarked form
\begin{eqnarray*}
p ( f^u | m_{\textsc{subj}} )
&=&
1
\\
p ( f^a | m_{\textsc{subj}} )
&=&
1 - p ( f^u | m_{\textsc{subj}} )
\ =\
0
\end{eqnarray*}
\end{exe}
Now consider a speaker expressing $m_{\textsc{obj}}$.  The speaker/learner  already knows how to express the object.  In fact they have two different ways to express it.  So they must choose between forms, which means learning how \textit{not} to express the object.  Which form, if any, should be avoided?  

Since the speaker/learner is assessing forms, they consider the histories (counts) of the forms, rather than the history of object expression per se.  They favor a form more if it is more likely to have the desired meaning (namely, \textsc{obj}) rather than other meanings (such as \textsc{subj}).  The top equation calculates that factor based on the count of prior uses of the form $f^u$ to express the object, as a fraction of the total number of times  $f^u$ expressed subject or object.  The marked \textsc{acc} form $f^a$ is the only other option, so their probabilities sum to one (recall \ref{onlyfufa}), as shown in the bottom equation.  

\begin{exe} \label{skeq}  
\ex Formula for expression of the object
\begin{eqnarray*} 
p ( f^u | m_{\textsc{obj}})
&=&
\frac{ c^u_{\textsc{obj}}  }{ c^u_{\textsc{subj}} + c^u_{\textsc{obj}}  }
\\ 
p ( f^a | m_{\textsc{obj}} )
&=&
1 - p ( f^u | m_{\textsc{obj}} )
\ =\
\frac{ c^u_{\textsc{subj}} }{ c^u_{\textsc{subj}} + c^u_{\textsc{obj}} }
\end{eqnarray*}
\end{exe}
The top equation is a simple HRE equation.  It differs from the original one in the Fundamental Model (Section \ref{fundy-sec}) in the absence of the parameter $\alpha$, our measure of the likelihood of syntax emerging, hence inappropriate for choosing between expressions within a mature grammar.  
The bottom equation is derived from the top one.  
Consequently the choice of form $f^a$ is seen to depend on data from form $f^u$.  
 This learning model is guiding speakers to avoid the more ambiguous form $f^u$, in favor of the more informative form $f^a$.  It is interesting that this effect follows directly from reinforcement learning itself; see the Section \ref{efficiency-sec} below for more discussion of informativity.

\subsection{Result (i): when the unmarked is more frequent}
\label{unmarked-freq-sec}
In the above example we assumed that subjects are more frequent than objects, hence:

\begin{exe} 
\ex \label{again} \ \ \ \ \ 
$p(m_{\textsc{subj}}) > p(m_{\textsc{obj}})$
\end{exe}
The frequency based learning interacts with this meaning bias, just as in the Fundamental Model.  
Interestingly, we get essentially the same result as in the Fundamental Theorem (Emergence of Semantic Composition), 
when $p(m_{\textsc{subj}}) >.5$ (greater than its only alternative), but we get a markedly 
different result when $p(m_{\textsc{subj}}) <.5$.  Here is the theorem for the former case:

\begin{thrm}  
{\bf Form Competition Model, p$\,>\,$1/2}

Suppose $p(m_{\textsc{subj}} ) > .5$ in the Form Competition 
production algorithm.
Then, for any start values $[c_{\textsc{subj}}^u]^0,\, [c_{\textsc{obj}}^u]^0 >0$, 
as the number of utterances $k$ in the language history 
grows we have as $k \to \infty$,


a) the probability a speaker with message $m_{\textsc{subj}}$  chooses form $f^u$  converges to 1,

b) the probability a speaker with  message $m_{\textsc{obj}}$ chooses form $f^u$ converges to 0.

c)  probability a hearer interprets  $f^u$ as $m_{\textsc{subj}}$ converges to 1,

d)  probability a hearer interprets  $f^u$ as $m_{\textsc{obj}}$ converges to  0.
\end{thrm}
See Appendix D for a proof.
Convergence to categoricalization is predicted for
$p(m_{\textsc{subj}}) >.5$, that is, if the unmarked meaning is more
common than the marked one for which a  form was recruited.  
This result is confirmed by two case studies from the history of English in Section \ref{dia-sec}.

\subsection{Result (ii) when the marked is more frequent}
\label{marked-freq-sec}
When $p(m_{\textsc{subj}}) <.5$, the language is predicted to settle at Stage II. 
\begin{exe} 
\ex \label{again}
$p(m_{\textsc{subj}}) < p(m_{\textsc{obj}})$
\end{exe}
The recruited form is predicted to rise in frequency and level off at
a frequency determined by the value of  $p(m_{\textsc{subj}})$.   This
means that the language settles at Stage II (see Table \ref{sotable}).

\begin{thrm} {\bf  Form Competition Model, p$\,<\,$1/2}

Suppose $p = p( m_{\sc subj} ) < .5$ in the Form
Competition production algorithm.  Then, for any start values
$\left[ c_{\sc subj}^u \right]^0 $, 
$\left[ c_{\sc obj}^u \right]^0 > 0$, 
as the number of utterances in the
language history grows, we have as $k \to \infty$,

a) the probability a speaker with message
$m_{\sc subj}$ chooses form $f^u$ is always 1,

b) the probability a speaker with message $m_{\sc obj}$ 
chooses form $f^u$ converges to $(1 - 2p) / ( 1 - p)$,

c) the probability a hearer interprets $f^u$ as
$m_{\sc subj}$ converges to $p / ( 1 - p)$,

d) the probability a hearer interprets $f^u$ as
$m_{\sc obj}$ converges to $( 1 - 2p)  / ( 1 - p)$.

\end{thrm}
See Appendix D for a proof.

\begin{table}\centering
\begin{tabular}{c|c|c}
$p(m_{\textsc{subj}})$    &    $f^u$-$m_{\textsc{obj}}$  &   $f^u$-$m_{\textsc{subj}}$  \\
\hline
$>0.5$   &      0                &                1   \\
\hline
0.49        &    .04           &        0.96  \\
0.4   &   1/3         &    2/3 \\
1/3   &   1/2       &     1/2 \\
0.25  &  2/3  &   1/3 \\
0.2    &  3/4  &  1/4 \\
0.1   &  8/9  &  1/9 \\
0.01  &  0.99 &  0.01  \\
\hline
0       &     1   &   0 \\
\hline
\end{tabular}
\caption{\label{lesstable}Sample values derived from Theorem 5.}
\end{table}
Table \ref{lesstable} give some sample values derived from the formulas in 
Theorem 5.  The first row represents the generalization emerging from Theorem 4 
discussed in the previous section (Sec. \ref{unmarked-freq-sec}).  
The other rows show predicted outcomes for cases where 
$m_{\textsc{obj}}$ is  more common than $m_{\textsc{subj}}$.  
For example, if one third of the tokens are $m_{\textsc{subj}}$ (and the other two thirds are $m_{\textsc{obj}}$), then the language is predicted to settle at a state where $f^u$ is  equally likely to express either of the two meanings, $m_{\textsc{subj}}$ and $m_{\textsc{obj}}$.  In other words, all of the rows in Table (\ref{lesstable}) except the top and bottom rows represent languages 
that do not progress to Stage III but rather settle at Stage II.  See Section \ref{efficiency-sec} below for related discussion.

\subsection{Objects expressed by word order}

We saw above how accusative case could emerge and distinguish objects from subjects.  In the absence of a case marker, word order could 
fulfill this function.  Since the subject is preverbal, the object must be post-verbal. So VO order plays the same role as accusative case above.  This is shown in Table \ref{ovvo}.

\begin{table}\centering 
\begin{tabular}{l|c|c}
&  NP V(P)   &  V NP \\
\hline
Stage I &  $\textsc{subj} \vee \textsc{obj} $  & \\
Stage II  &  $\textsc{subj}  \vee \textsc{obj} $ &  $\textsc{obj} $ \\
Stage III  &  $\textsc{subj} $ &   $\textsc{obj} $ \\
\end{tabular}
\caption{\label{ovvo} From SOV to SVO word order. }
\end{table}

\subsection{Case studies}
\label{dia-sec}

The Fundamental Model accounts for the emergence of syntax itself, and so it is difficult to find good data to test the model, although this could be an area of future research.  
But data can be found for testing the Form Competition Model, because it starts from a mature language that has converged on messages to be expressed by its grammatical relations, and needs to settle on forms for its grammatical relation.   We illustrate the Form Competition Model with two case studies, the rise of pronoun obviation \citep{mattausch2005,keenan:2008} and  the rise of the imperfective-progressive split \citep{deo2015}.

\subsubsection{Case study 1: Anaphoric binding}
\label{anaphoric-sec}
Our first example is from  anaphoric binding in English, which currently exhibits the pattern shown in \ref{mary}.   The \textit{self}-forms such as \textit{herself} require a local antecedent, while ordinary pronouns such as \textit{her} reject a local antecedent:

\begin{exe}
\ex\label{mary}
\begin{xlist}
\ex Mary_i admires herself_i.  \hfill \textit{conjoint} 
\ex Mary_i admires her_{*i}.   \hfill \textit{disjoint} 
\end{xlist}
\end{exe}
The \textit{self}-forms are said to have the \textit{conjoint} reading, while the ordinary pronouns have the \textit{disjoint} reading.  This has been the case in English since the late 1700s and through to the present day.  
This period exemplifies Stage III in Table \ref{progtable}, where the pronoun is $f^{\textsc{pron}}$ and the \textit{self}-form is $f^{\textsc{self}}$.  
 Let us consider the history of English leading up to this state.   

\begin{table} \centering
\begin{tabular}{|l|c|c|}
\hline
&  $f^{\textsc{pron}}$   &  $f^{\textsc{self}}$\\
\hline
Stage I &  \textit{disjoint} $\vee$ \textit{conjoint}  & \\
Stage II  &  \textit{disjoint} $\vee$ \textit{conjoint} &  \textit{conjoint}  \\
Stage III  &  \textit{disjoint} &   \textit{conjoint}  \\
\hline
\end{tabular}
\, \, 
\begin{tabular}{|l|c|c|}
\hline
&  $f^{\textsc{imp}}$   &  $f^{\textsc{prog}}$\\
\hline
Stage I &  \textit{struct.} $\vee$ \textit{phen.}  & \\
Stage II  &  \textit{struct.} $\vee$ \textit{phen.} &  \textit{phen.}  \\
Stage III  &  \textit{struct.} &   \textit{phen.}  \\
\hline
\end{tabular}
\caption{\label{progtable}Recruitment and categoricalization.}
\end{table}

In Stage I the pronoun forms were used in both disjoint and conjoint contexts:

 \begin{exe}
\ex\label{beowulf} 
\gll sy\dh\TH {an} he hine to gu\dh{e} gegyred h\ae fde \\
once he.\textsc{nom}_i him.\textsc{acc}_i   for battle girded had  \\ 
    \glt `once he had girded himself (lit. `him') for battle'  (c750, Beowulf 1473) 
\end{exe}
In a usage still found today, the \textit{self}-forms were sometimes used  appositionally as markers of surprisal or contrast:

 \begin{exe}
\ex\label{christ} 
\gll hw\ae t Crist self t\oe hte and his apostolas on {\TH \oe re} niwan gecy\dh nisse, \dots \\
what Christ self.\textsc{nom.sg} taught and his apostles  in the New Testament, \ldots  \\ 
    \glt `what Christ himself and his apostles taught in the New Testament'  (c1000; \AE O \& N Pref)
\end{exe}
The \textit{self}-forms in object positions were subsequently recruited for conjoint readings, bringing us to Stage II:

 \begin{exe}
\ex\label{middle} 
 \gll  {\TH e} {tre[sur]} {\TH at} godd {\textyogh ef} him seolf fore \\
 the treasure that God_i gave [him self]_i for \\
    \glt `the treasure that God gave himself for'  (c1200; Sawles Warde)
\end{exe}
However, conjoint readings of ordinary pronouns, as in \ref{beowulf}, persisted during this stage, so the two forms were alternating.  
This set the stage for categoricalization, which we analyze in what follows.

Table \ref{keenan} gives the historical progression of frequency counts in conjoint readings, from a corpus study by \citet{keenan:2008}. In Old English corpora, 18\% of conjoint object pronouns are \textit{self}-forms while the remaining conjoint cases are bare pronouns.  During this Stage II, the share of the conjoint contexts with \textit{self}-forms grew larger and the share of pronouns smaller, a process that accelerated greatly in the 1500s.  By the 1700s there were very few conjoint readings of direct object bare pronouns.   In contrast to direct objects, objects of prepositions are not quite as local to a subject antecedent, especially if the preposition introduces its own predicate, which could explain why unmarked objects of semantically rich prepositions sometimes allow conjoint binding, as in \textit{John_i tossed the can behind him_{i/j}}.  

\begin{table}\centering
\begin{tabular}{l|llllll}
 &  \multicolumn{3}{c}{\underline{Object of V}} & 
 \multicolumn{3}{c}{\underline{Object of P}}\\
 & Pron &  Self &  \%Self  & Pron &  Self &  \%Self \\
 \hline
c750-1154 & 419 & 89  & 18\%  & 96 &  21  & 18\% \\
1154-1303 & 735 & 167 & 19\%  & 159 &  102   & 39\%  \\
1303-1400 & 753 & 191 & 20\%  & 131 &  122   & 48\% \\
1400-1495  & 915  & 203 & 18\%  & 121 &  55   & 31\% \\
1495-1605  & 291 &  1232 & 81\% &  167 &  291   & 64\% \\ 
1605-1700  & 138  & 930 & 87\%  & 162  & 336   & 67\%  \\
1722-1777 & 3 &  335 & 99\%  & 28  & 73  & 72\% 
\end{tabular}
\caption{\label{keenan} English pronouns and \textit{self}-forms with local antecedents.  Data from \cite{keenan:2008}.}
\end{table}

The rise of the use of self-forms to express conjoint readings is predicted from the Form Competition Model theorem above (Section \ref{marked-sec}), but only if disjoint readings are more frequent than conjoint readings overall.  In fact that appears almost certainly to be the case.  
Table \ref{cocatable} provides an estimate of the relative frequency of disjoint versus conjoint objects in English, based on counts of object pronouns (in accusative case) immediately following a verb.  Over 90\% are disjoint.  
These data are from recent corpora only, but there is no reason to expect earlier stages of the language to have a very different distribution.  

\
\begin{table}\centering
\begin{tabular}{l|ccrrr}
	&	\%pron	&	\%\textit{self}-form	&	pron	& \textit{self}-form & total \\
	\hline 
1sg	&	94.5		&	5.5		&	1,654,131 &	96,796 & 1,750,927 \\
1pl	&	94.6		&	5.4		&	558,068 &  32,147 & 590,215 \\
3msg	&	89.8		&	10.2		&	1,109,058  & 126,166 &  1,235,224 \\
3fsg	&	90.9		&	9.1		&	599,693 & 59,864 & 659,557\\ 
3pl	&	90.7		&	9.3		&	1,011,373 & 103,303 & 1,114,676\\
\end{tabular}
\caption{\label{cocatable} Frequency of object pronouns and reflexives in the Corpus of Contemporary American English (www.english-corpora.org/coca). 
Words immediately following a verb, including pronouns \textit{me, us, him, her, them} and \textit{self}-forms \textit{myself, ourselves, himself, herself, themselves}.}
\end{table}

To get an intuition for how the account works, consider a speaker of English during Stage II, for example in the 1500s, when the \textit{self}-forms were on the rise in conjoint contexts.  The speaker has a conjoint message such as \ref{mary}a in mind, and she is considering the use of the bare pronoun \textit{her}.  She counts up previous tokens of \textit{her} in object position that she has witnessed, and determines what shares of the total were conjoint versus disjoint uses.  A constant exogenous factor is affecting the frequency of the input: disjoint messages are more frequent than conjoint ones overall (see Table \ref{cocatable}).  This depresses the conjoint count, making the speaker unlikely to use the form for a conjoint meaning, which further depresses that count for future speakers, and over time they drop to near zero, at which point the language has reached Stage III.

\subsubsection{Case study 2: The rise of the progressive}
\label{prog-sec}
Our second application of the Form Competition Model is the rise of the English periphrastic progressive \textit{BE V-ing} form shown in (\ref{jane}b):

\begin{exe}
\ex\label{jane}
\begin{xlist}
\ex Jane sorts the mail.  \hfill \textit{imperfective}
\ex Jane is sorting the mail.   \hfill \textit{progressive}
\end{xlist}
\end{exe}
\textsc{Imperfective} and \textsc{progressive} are our names for the verb forms such as \textit{sorts} (\ref{jane}a) and \textit{is sorting} (\ref{jane}b), respectively.  
The meanings expressed by these forms in modern English will be called 
\textsc{structural} and \textsc{phenomenal}, following \citet{deo2015}, who follows \citet{goldsmith1982logic}.  
Sentence (\ref{jane}a) tells us about the \textit{structure} of the world, namely that Jane generally sorts the mail.     Sentence (\ref{jane}b) refers to a specific \textit{phenomenon}, an episode of mail-sorting.  
In modern English sentences in present tense and with present time reference, the imperfective is used for structural and progressive for phenomenal reference.  Statives such a \textit{Jane likes Mary} are structural, as are habitual or iterative sentences like (\ref{jane}a), and imperfective is used for all of these.  Episodic event readings are expressed with the progressive, when in present tense.

The stages of development are shown in Table \ref{progtable}.  Old English  at the beginning of the 1400s was in Stage I:  as in many languages a single verb form, the imperfective, was used for both structural and phenomenal judgments.  During the 1400s the progressive form was recruited (from adjectives in \textit{-ing}) for phenomenal reference, bringing English into the start of Stage II.  The progressive grew more frequent, until it displaced the imperfective for the expression of phenomenal judgments, bringing us to Stage III, where English remains today.   

It remains to be shown that structural judgments outnumber phenomenal
ones overall.  Once again we
can use frequency counts from contemporary English to estimate this,
since we are in Stage III and the forms line up with the meanings.
Those estimates are presented in Table \ref{cocaaspect}.\footnote{Our
  heuristic for simple present was to count anything that the NLTK part of speech
  tags as a VBZ (Penn tagset, 3rd person singular present verb form)
  except \textit{is}.  This undercounts as it omits copular
  constructions. Our heuristic for progressive is to count two directly adjacent words, the first being \textit{is} or \textit{are}, the second having part of speech tag VBG (present participle \textit{V-ing} form).  This also undercounts as it omits those clauses with a word intervening between \textit{is} and the VBG.}
As shown from the percentages in the first two columns, the structural judgments greatly outnumber phenomenal ones.

\begin{table}\centering
\begin{tabular}{l|rrrrr}
genre	&	\%impf.	&	\%prog.	&	impf.	& prog. & total \\
	\hline 
fiction &     93.9             &     6.1       &   985,346  &   63,722  &   1,049,068        \\ 
spoken &     82.7           &     17.3       &  1,072,419    &   224,624    &   1,297,043             
\end{tabular}
\caption{\label{cocaaspect} Frequency of imperfective and progressive sentences in the Corpus of Contemporary American English (www.english-corpora.org/coca). }
\end{table}

\subsection{The emergence of efficiency}
\label{efficiency-sec}

The Form Competition Model predicts that certain grammars will emerge (Section \ref{unmarked-freq-sec}), while others will not (Section \ref{marked-freq-sec}).  
As it turns out, the predicted grammars are more efficient for communication than the unpredicted grammars, an interesting result in light of the plethora of recent work demonstrating the efficiency of human languages \citep[, among others]{Piantadosi+etal:2011,Gibson+etal:2019, mollica2021forms,chen+etal:2023}.   
Communicative efficiency has an intuitive definition: `a code is efficient if successful communication can be achieved with minimal effort on average by the sender and receiver, usually by minimizing the message length.' \citep[390]{Gibson+etal:2019}.  
Efficiency has been quantified  as an optimal balance between \textsc{informativeness} and \textsc{complexity} of the signal.

\textsc{Informativeness}, a key concept from  Shannon's (1948) \nocite{Shannon:1948} theory of communication, is known by other terms such as surprisal, (un)predictability, and conditional Shannon entropy.  The informativeness of a sign is defined as the log of the inverse of its probability, given its local context.  The higher the probability, the lower the informativeness; as a consequence, an unambiguous form in context is more informative than an ambiguous one in the same context.  Formal \textsc{complexity} is a measure of code length and can be defined in various ways, such as the count of phonemic segments or morphemes in the sign.  Efficiency involves the correlation between those two measures: in an efficient system more complex forms are more informative (= more surprising = less predictable) in their contexts, while less complex forms are less informative (= less surprising = more predictable) in their contexts.  One important example of human language efficiency is that shorter words are found in more predictive contexts overall \citep{Piantadosi+etal:2011}.   
  
The Form Competition Model is charged with choosing between competing forms, such as \textit{her} versus \textit{herself} in the pronoun obviation case above (Section \ref{anaphoric-sec}).  The marked form \textit{herself} is the longer of the two, with two morphemes instead of one, and more phonological segments.  Notice that in object position \textit{herself} is also more informative than \textit{her}.  The reason is that the conjoint (reflexive) meaning is rarer than the disjoint, according to our message probabilities.   Hence \textit{herself} is less predictable, or more informative.  So the greater code length correlates with greater informativity; English is efficient, in that respect.  

Imagine a language like English, except that Stage II proceeded differently: instead of \textit{-self} being recruited to mark the conjoint pronoun use,  the  morpheme \textit{-other} had been recruited to mark the disjoint uses.  The language at Stage III would be identical to English except that English \textit{her} is replaced with \textit{her-other}, and English \textit{herself} is replaced with \textit{her}.  Such a language would be a notational variant of English with exactly the same expressive capabilities, but it would be less efficient than English. The reason is that the longer form \textit{her-other} in object position, is less informative than the shorter form \textit{her} in object position.  The point here is that such an inefficient language will not emerge under the Form Competition Model.  Instead the model predicts that a disjoint marker \textit{-other} would continue to be optional, settling at a rate determined by overall probability of disjoint versus conjoint messages.  

In sum, an extra morpheme is often recruited to disambiguate a construction.  It can mark either the more common or the rarer of the two meanings.  A language is more efficient if it marks the rarer meaning, and such an efficient language is predicted to converge on a categorical rule, under the Form Competition Model.

\section{Conclusion}

In this paper we have shown how a grammatical system can emerge from a set of grammatically unstructured words.  Apart from a well-established reinforcement learning mechanism, the most crucial model assumption is that people have general preferences about what to talk about, given a rather specific utterance context.  

The paper stands as a proof of concept: we have literally provided proofs that certain grammar systems will emerge, given the assumptions of the models.  Numerical simulations demonstrate the plausibility of the theory.  Only a few important grammatical structures are illustrated in this paper, but the models appear to have wide applicability.  The Fundamental Model and its extensions account for the emergence of  semantic composition rules for combining two signs.  The Form Competition Model accounts for the emergence of a grammatical system of oppositions, such as a two case system with  nominative and accusative forms distinguishing the subject and object, respectively.  But the larger goals of this paper are to inspire further work in two areas.  In recent decades the quantitative analysis of usage data has yielded exciting results, bringing us closer to solving the mysteries of human language.   We have proposed a particular type of theoretical foundation for such usage-based research on natural language.  
Our theory implies  a method of analysis, which we have illustrated using small artificial languages and two case studies on the history of English.  
   Second, we hope to inspire further development of that foundation, whether it stays within reinforcement learning, or incorporates other approaches.

\bibliography{emergence}

\appendix
\appendixpage
\section{The Fundamental Model: theorems and proofs}

\subsection{The Fundamental Theorem for Speakers}

Language histories' state of learning 
${\mathbf c}^k = (c_1^k,c_2^k,\ldots,c_I^k)$ 
is a Markov process $\{\mathbf c^k\}_{k=0}^\infty$  in the state
space $\mathbb{R}^I_+$
on a probability space $(\Omega, {\cal F}, P)$ where
each $\omega \in \Omega$  corresponds to a single 
language history 
(in this appendix
we use $P$ instead of $p$ to denote the probability). 
This probability space has a countable set of independent random
variables $\xi_{ik}$,  $i=1,2,\ldots,I'$, $k = 1,2,\ldots$,
with uniform distribution on $[0,1]$. ${\cal F}_k$ is
the $\sigma$-algebra generated by 
$\{ \xi_{i,k'} \}_{ k' \leq k, i \geq 1 }$.
The random variables along with the message probabilities
and HRE formulas determine the choices made 
in the language production algorithms, hence define 
the Markov process $\{\mathbf c^k\}_{k=0}^\infty$. In particular,
$\xi_{1,k}, \xi_{2,k}, \ldots, \xi_{I',k}$ determine the 
choices made when producing the $k$th utterance in 
a language history. 

The Fundamental Theorem and its proof is
similar to results and proofs of theorems 1 and 2 in 
\cite{beggs2005convergence},
which are 
a generalization of a result in 
\cite{pemantle1999vertex}. 
Beggs makes some different assumptions
(e.g., Beggs assumes positive payoffs
at every $k$). 

\begin{thrm} {\bf (The Fundamental Theorem for Speakers: 
Emergence of Semantic Composition)} 

Suppose the production 
algorithm in 3.3 has messages $m_1, m_2, \ldots, m_I$ and
$P(m_1) > P(m_i)$ for $i \neq 1$. Then for any values
$\alpha > 0$ and $c_1^0, \ldots, c_I^0 > 0 $, as the number
of utterances $k$ in the language history grows we have
for $i \neq 1$

a) the count ratio $c_i /c_1 $ converges to 0,

b) the probability a speaker chooses a phrasal utterance for 
message $m_1$ converges to 1,

c) the probability a speaker chooses a phrasal utterance for 
message $m_i$ converges to 0.
\end{thrm} 
{\bf Proof: } Lemma 2 given below proves  
statement a). Statement b) follows from statement a), 
Lemma 1, and the HRE equation below with $j=1$. 
Statement c) similarly follows with $j=i$. 
\begin{align*}
P \left( u^* | m_j \right)
\ =\
\frac{ c_j }{ c_1 + c_2 + \ldots + c_I + \alpha }
&\ =\
\frac{ c_j / c_1  }{ 1 + c_2 / c_1 + \ldots + c_I / c_1  + \alpha / c_1 }
\end{align*}
$\blacksquare$

\begin{lem} In the Fundamental Model, if 
$c_i^0 > 0$ and $P(m_i|s) > 0$, then $c_i^k \to \infty$ 
almost surely as $k \to \infty$.
\end{lem}
{\bf Proof: }
Let 
\begin{align*}
R^k_i
&\ =\
\frac{ c_i }{ c_1 + \ldots + c_I + \alpha },
\ \ \ \ 
\ \ \ \ 
\ \ \ \ 
p_i
\ =\
P(m_i | s ).
\end{align*}
For any $k \geq k^* =
 c_1^0 + \ldots + c_I^0 + \alpha$ we take
the conditional expected value of $1/c_i^{k+1}$. 
\begin{align*}
\mbox{E}\left[ \left. \frac{ 1 }{ c^{k+1}_i } \right| {\mathbf c} 
= {\mathbf c}^k  
\right]
&\ =\
\frac{ 1 }{ c_i + 1 }  R_i p_i  
+
\frac{ 1 }{ c_i  } ( 1 -  R_i p_i )
\\
&\ =\
\frac{ 1 }{ c_i }
\left[ 
1
- 
\frac{ 1 }{ c_i + 1 } 
\right]   R_i p_i
+
\frac{ 1 }{ c_i  } \left( 1 -   R_i p_i \right)
\\
&\ \leq\
\frac{ 1 }{ c_i  } 
\left[
1
- 
\frac{ c_i }{ c_i + 1 } 
\frac{ 1 }{ 2 k } p_i
\right]
\\
&\ \leq\
\frac{ 1 }{ c_i  } 
\left[
1
- 
\frac{ C }{ k } 
\right]
\ \ \ \ \
\ \ \ \ \
C
\ =\
\frac{ c_i^0 }{ 2 (c_i^0 + 1) } 
\end{align*}
Taking expectations we get 
\begin{align*}
\mbox{E}\left[ 
\frac{ 1 }{ c^{K+1}_i } 
\right]
&\ =\
\mbox{E}\left[ 
\mbox{E}\left[ 
\left. \frac{ 1 }{ c^{K+1}_i } \right| {\mathbf c}^K 
\right]
\right]
\\
&\ \leq\
\mbox{E}\left[ \frac{ 1 }{ c^{K}_i } 
\right]\left[
1
- 
\frac{ C }{ K } 
\right]
\\
&\ \leq\
\mbox{E}\left[ \frac{ 1 }{ c_i^{k^*} }
\right]
\prod_{k = k^*}^K
\left[
1
- 
\frac{ C }{ k } 
\right]
\\
&\ \leq\
\frac{ 1 }{ c_i^0 }
\prod_{k = k^*}^K
\left[
1
- 
\frac{ C }{ k } 
\right]
\ \to\ 0
\ \ \ \ 
\mbox{ as } \
K \to \infty.
\end{align*}
Thus $1/c_i^k$ is a positive and bounded supermartingale.
Using Doob's martingale convergence theorem, $1/c_i^k \to 0$
almost surely. 
$\blacksquare$ 
\begin{lem} If $ p_1 > p_i$, then $c_i^k / c_1^k \to 0$ as 
$k \to \infty$, almost surely. 
\end{lem} 
{\bf Proof: } 
For any $\delta >0$ and since $c_1^k \to \infty$ almost surely, 
there is a $k^* \geq c_1^0 + \ldots + c_I^0 + \alpha $ 
and a set $A \in {\cal F}_{k^*} $ with 
$P(A) > 1 - \delta$  such that for all $k \geq k^*$
\begin{align*}
p_1 \frac{ c^k_1 }{ c^k_1 + 1 } - p_i 
&\ \geq\
\frac{ p_1 - p_i }{ 2}
\ =\ 
C > 0
\ \ \ \ \
\mbox{ on } \
A.
\end{align*}
Let $1_A$ be the indicator set of $A \subset \Omega$ 
where $1_A(\omega) =1$ for 
$\omega \in A$ and 0 otherwise.
\begin{align*}
\mbox{E}\left[ \left. 1_A \frac{ c^{k+1}_i }{ c^{k+1}_1 } \right| {\mathbf c}
=   {\mathbf c}^k \right]
&\ =\
1_A \left[ 
\frac{ c_i }{ c_1 + 1 } 
R_1 p_1 
+
\frac{ c_i + 1 }{ c_1 } 
R_i p_i 
+
\frac{ c_i }{ c_1  } ( 1 -  R_1 p_1 -  R_i p_i )
\right]
\\
&\ \leq\
1_A \left[ 
\frac{ c_i }{ c_1} 
\left( 
1 
- 
\frac{ 1  }{ c_1 + 1 }
\right)R_1 p_1 
+
\frac{ 1 }{ c_1 } 
R_i p_i 
+
\frac{ c_i }{ c_1  } ( 1 -  R_1 p_1 )
\right]
\\
&\ =\
1_A  
\frac{ c_i }{ c_1} 
\left( 
1 
+ 
\frac{ 1 } { c_1 + \ldots + c_I + \alpha }
\left(
p_i
-
p_1 
\frac{ c_1 }{ c_1 + 1 }
\right)
\right)
\\
&\ \leq\
1_A 
\frac{ c_i }{ c_1} 
\left( 
1 
- 
\frac{ C } { 2 k }
\right)
\end{align*}
Taking expectations we get 
\begin{align*}
\mbox{E}\left[ 
1_A \frac{ c^{K+1}_i }{ c^{K+1}_1 } 
\right]
&\ =\
\mbox{E}\left[ 
\mbox{E}\left[ 1_A 
\left. \frac{ 1 }{ c^{K+1}_i } \right| {\mathbf c}^K 
\right]
\right]
\\
&\ \leq\
\mbox{E}\left[ 1_A  \frac{ c^{K}_i }{ c^{K}_1 } 
\right]\left[
1
- 
\frac{ 2 C }{ K } 
\right]
\\
&\ \leq\
\mbox{E}\left[ 1_A \frac{ c_i^{k^*} }{ c_1^{k^*} }
\right]
\prod_{k = k^*}^K
\left[
1
- 
\frac{ 2 C }{ k } 
\right]
\\
&\ \leq\
\frac{ 2k^* }{ c_1^0 }
\prod_{k = k^*}^K
\left[
1
- 
\frac{ 2 C }{ k } 
\right]
\ \to\ 0
\ \ \ \ 
\mbox{ as } \
K \to \infty.
\end{align*}
Thus $1_A c_i^k/c_1^k$ is a positive 
and bounded supermartingale.
Using Doob's martingale convergence theorem, 
$1_A c_i^k/c_1^k \to 0$ almost surely.  Since
$\delta>0$ was arbitrary, we get almost sure convergence
on $\Omega$. 
$\blacksquare$

\subsection{The Fundamental Theorem for Hearers}

In (\ref{BayesPosterior}) Bayes's Rule is used to estimate the probability 
that the utterance 
[{\it Cat walk.}] ($u^\star$), has the message mapping
$m_{\textsc{walker}}$ ($= m_1$).

\begin{exe}\ex \label{BayesPosterior}
\begin{align*}
p(m_1, s| u^\star )
&=
\frac{ p( u^\star | m_1, s) p(m_1 | s) p(s) }{ p(u^\star ) }
\\
&=
\frac{ p( u^\star | m_1, s) p(m_1 | s) }{ p(u^\star | m_1, s) p(m_1 | s) +
p(u^\star | m_2, s) p(m_2 | s)  }
\\
&=
\frac{c_1  p(m_1 | s) }{ c_1 p(m_1 | s) +
c_2 p(m_2 | s)  }
\ =\
\frac{  p(m_1 | s) }{ p(m_1 | s) +
(c_2/c_1) p(m_2 | s)  }
\end{align*}
\end{exe}

Similarly  
\begin{exe}\ex \label{BayesPosterior2}
\begin{align*}
p(m_2, s | u^\star )
&=
(c_2/ c_1)\frac{   p(m_2 | s ) }{ p(m_1 | s ) +
(c_2/c_1) p(m_2 | s )  }.
\end{align*}
\end{exe}
Using the theorem  from the previous section
and assuming as before that $p(m_1|s) > p(m_2|s)$, 
we saw that $c_2/c_1$ converges to zero.
Thus it also follows that $p( m_1, s | u^\star )$ converges 
to 1 and $p( m_2, s | u^\star )$ converges  to 0 
 (from \ref{BayesPosterior} and \ref{BayesPosterior2}, respectively). 
 Hence the hearer learns the syntax 
and the semantic composition rule, and knows that a single two word phrase of the form
{\it Cat walk.} means that the cat is the walker.  

Thus we have the second aspect of the emergence
of semantics.

\begin{thrm} {\bf The Fundamental Theorem for Hearers: 
the Emergence of Semantic Interpretation}

Suppose the production 
algorithm in 3.3 has messages $m_1, m_2, \ldots, m_I$ and
$P(m_1) > P(m_i) $ for $i \neq 1$. 
Then for any values
$\alpha > 0$ and $c_1^0, \ldots, c_I^0 > 0 $, as the number
of utterances $k$ in the language history grows we have
for $i \neq 1$

a) the count ratio $c_i /c_1 $ converges to 0,

b) the probability a hearer interprets a phrasal utterance as message $m_1$ converges to 1,

c) the probability a hearer interprets a phrasal utterance as message $m_i$ converges to 0.
\end{thrm}
{\bf Proof: } Statement a) follows from the Fundamental Theorem.  We use Bayes to prove b) and c).
\begin{align*}
P \left( m_1 | u^* \right)
&\ =\
\frac{ P \left( u^* | m_1  \right) P \left(m_1 \right) }
{ P \left( u^* \right) }
\ =\
\frac{ P \left( u^* | m_1  \right) P \left( m_1 \right) }
{ \sum_{i' = 1}^I P \left( u^*  | m_{i'} \right)  P \left( m_{i'} \right) }
\\
&\ =\
\frac{ c_1 P \left( m_1 \right) }
{ \sum_{i' = 1}^I  c_{i'}  P \left( m_{i'} \right) }
\ =\
\frac{  P \left( m_1 \right) }
{ \sum_{i' = 1}^I  c_{i'}/c_1  P \left( m_{i'} \right) }
\ \to\
1
\end{align*}

\begin{align*}
P \left( m_i | u^* \right)
&\ =\
\frac{ P \left( u^* | m_i  \right) P \left(m_i \right) }
{ P \left( u^* \right) }
\ =\
\frac{ P \left( u^* | m_i  \right) P \left( m_i \right) }
{ \sum_{i' = 1}^I P \left( u^*  | m_{i'} \right)  P \left( m_{i'} \right) }
\\
&\ =\
\frac{ c_i / c_1  P \left( m_i \right) }
{ \sum_{i' = 1}^I  c_{i'} / c_1 P \left( m_{i'} \right) }
\ \to\
0
\end{align*}
$\blacksquare$


\section{A production algorithm for the Sequential Model}

\noindent
\textbf{Step 1. Select a message.}  
Given a drinking scene $s^k_{\textsc{drink}}$ of a cat drinking milk,  
the speaker selects a message from the following probability distribution:
\begin{exe} \ex \label{walkprob2}
$1
= 
p(m_{\textsc{drinker}} | s_{\textsc{drink}} ) +  
p(m_{\textsc{drinkee}}  | s_{\textsc{drink}}  )  $
\end{exe}  

The speaker selects a message, either 
$m_1 = m_{\textsc{drinker}}$ (`a cat is drinking (something).')
or 
$m_2 = m_{\textsc{drinkee}}$ (`(something)  is drinking milk.').  
We will
use index $g$ to iterate over grammatical relations. 
At first, {\bf we set $g$ = 1}, so that the first grammatical relation 
to be considered is $GR_1 = \textsc{subj}$.

\medskip

\noindent
{\bf Step 2. Produce an utterance.} The speaker considers
uttering the two-word phrase $u^*_{GR_g}$ with the grammatical 
relation $GR_g$. She uses the HRE formula (\ref{gthhre}) for the $g$th 
grammatical relation $GR_g$, from the sequence $g=1,2$ of 
HRE formulas (10) in this model. The two words in the uttered  
phrase $u^*_{GR_g}$ are {\it drink} and the word for 
the individual 
filling the role in the message, either {\it cat} or {\it milk}. 

Each grammatical relation
has its own state of learning ${\bold c}_g = (c_{1g}, c_{2g})  $
where $c_{ig}=c_{ig}^{k-1}$ equals the sum of
the starting value $c_{ig}^0$ plus the number of phrasal
utterances $u^*_{GR_g}$ expressing message $m_i$ among the 
previous $k-1$ utterances. 
Then, for the current value 
of $g$, the probability the $k$th 
utterance with message $m_i$ is the phrase $u^*_{GR_g}$
is given by equation (\ref{gthhre}) with parameter $\alpha_g \geq 0$.

\begin{exe}\ex\label{gthhre}
$p(u^\star_{GR_g}| m_i)
= 
\displaystyle \frac{ c_{ig} }{ c_{1g} + c_{2g} + \alpha_g } $
\ \ \ \ \
\ \ \ \ 
\ \ \ \ 
\mbox{Harley-Roth-Erev formula for message m_i, $g$th GR}
\end{exe}
If the speaker utters $u^*_{GR_g}$ for current value of
$g$, then continue to step 3,  otherwise, 
{\bf they increase the value of $g$ by 1 and return to the 
beginning of 
step 2}.  
 If all grammatical relations in the sequence 
fail to produce a phrasal sign $u^*_{GR_g}$, 
then leave all states of learning 
${\bold c}_g$ unchanged and return to step 1.

\medskip

\noindent
\textbf{Step 3. Update the history.} Update the phrasal utterance counts $c_{ig}$  
to reflect the outcome in Step 2 and return to Step 1.
More precisely, keep all ${\bold c}_g$ 
unchanged unless there is a phrasal sign  $u^*_{GR_g}$ in 
Step 2.  In that case,
add 1 to $c_{ig}$ and keep all $c_{i'g'}$ with $(i',g') \neq (i,g)$ unchanged
and then return to Step 1.

\section{The Model with Forms: 
theorem and proof}

\begin{thrm} {\bf Model with Forms (multiple messages and 
forms)}

Suppose we have messages $m_1,m_2,\ldots,m_I$ 
and forms $f_1,f_2, \ldots, f_J$ with probabilities
$P(m_i)$, $P(f_j | m_i ) $.  Let $c^k_{ij}$ be the count
of the number of times $j$th form with message $m_i$
is used (plus start value $c_{ij}^0>0$) after 
$k$ iterations of the 
Model with Forms production algorithm.  

If $P(f_1|m_1)P(m_1) > P(f_j|m_i) P(m_i)$ for all pairs
$(i,j) \neq (1,1)$ then the language history converges
to using form $f_1$ with message $m_1$.  

a) $c_{ij}/c_{11} \to 0$ for all pairs $(i,j) \neq (1,1)$, 

b) the probability a speaker with message $m_1$ 
chooses form $f_1$ converges to 1,

c) the probability a speaker with message $m_1$ 
chooses form $f_j$, $j \neq 1$, converges to 0.

d) the probability a speaker with message $m_i$,  
$i \neq 1$, chooses form $f_1$ converges to 0.

e) the probability a hearer interprets form $f_1$ as $m_1$
converges to 1,

f) the probability a hearer interprets form $f_j$,
$j \neq 1$ as $m_1$ converges to 0,

\end{thrm}
{\bf Proof: } This has the same structure as the Fundamental
Model. Define a Fundamental Model with 
messages $m'_1,m'_2, \ldots, m'_{N}$, $N = I\cdot J$
having probabilities given by $P(m'_n) = P(f_j | m_i) P(m_i)$,
$n = n(i,j) = i + (j-1)J$.  Then the counts $c'_{n(i,j)} = c_{ij}$,
which are the counts
in the Model with Forms. Since $P(m'_1) > P(m'_n)$ for 
$n \neq 1$, we have  by Fundamental Theorem 
$c_{ij}/c_{11} = c'_{n(i,j)}/c'_1 \to 0 $ for all 
$n(i,j) \neq 1$)  and $(i,j) \neq (1,1)$.  
Statemenst b), c) ... f) follow from statement a) as in the
Fundamental Theorems.
$\blacksquare$

\section{The Form Competition Model, $p\,>1/2$ and $p\,<1/2$:
theorems and proofs}

We use the following notation:
$m_1 = m_{\textsc{subj}} $, $m_2 = m_{\textsc{obj}} $,
$ c_1 = c_{\textsc{subj}} $, $c_2 = c_{\textsc{obj}}$,
$p = P( f^u |  m_{\textsc{subj}} )$. Then from (41), (42) in
`The Emergence of Grammar through Reinforcement Learning' we have
\begin{align*}
P( f^u  | m_1 ) 
&\ =\
1,
\ \ \ \ \
\ \ \ \ \
P( f^u | m_2 ) 
\ =\ 
\frac{c_2 }{  c_1 + c_2 },
\\
P( f^a  | m_1 ) 
&\ =\
0,
\ \ \ \ \
\ \ \ \ \
P( f^a | m_2 ) 
\ =\ 
\frac{c_1 }{  c_1 + c_2 }.
\end{align*}
\begin{lem} In the Form Competition Model, if  $c_i^0 > 0$ then 
$c^k_i \to \infty$ almost surely as $k \to \infty$.
\end{lem}
{\bf Proof: } The proof for $c_2$ is the same as in Lemma 1.
The proof for $c_1$ is the same as in Lemma 1, except
one derives the following inequality. 
\begin{align*}
\mbox{E} \left[ \left. 
\frac{ 1 }{ c_1^{k+1} } 
\right| (c_1,c_2) = \left( c^k_1, c^k_2 \right)  
\right]
&\ =\
\frac{ 1 }{ c_1 + 1 } p
+
\frac{ 1 }{ c_1  }( 1 -  p )
\\
&\ \leq\
\frac{ 1 }{ c_1 } 
\left[
1
-
\frac{ 1 }{ k + c_1^0 + 1 } p
\right]
\end{align*}
$\blacksquare$

\begin{thrm}  
{\bf Form Competition Model, p$\,>\,$1/2}

Suppose $p(m_{\textsc{subj}} ) > .5$ in the Form Competition 
production algorithm.
Then, for any start values $[c_{\textsc{subj}}^u]^0,\, [c_{\textsc{obj}}^u]^0 >0$, 
as the number of utterances $k$ in the language history 
grows we have as $k \to \infty$,


a) the probability a speaker with message $m_{\textsc{subj}}$  chooses form $f^u$  converges to 1,

b) the probability a speaker with  message $m_{\textsc{obj}}$ chooses form $f^u$ converges to 0.

c)  probability a hearer interprets  $f^u$ as $m_{\textsc{subj}}$ converges to 1,

d)  probability a hearer interprets  $f^u$ as $m_{\textsc{obj}}$ converges to  0.
\end{thrm}
{\bf Proof: } 
The proof is very similar to that of the Fundamental 
Theorem. 
For any $\delta >0$ and since $c_1^k \to \infty$ almost surely, 
there is a $k^*$ 
and a set $A \in {\cal F}_{k^*} $ with 
$P(A) > 1 - \delta$  such that for all $k \geq k^*$ we have 
$c_1 \geq 1$ on $A$. Let $1_A$ be the indicator function of 
$A$ and let $k \geq k^*$. 
\begin{align*}
&
\mbox{E} \left[ \left. 1_A
\frac{ c^{k+1}_2 }{ c_1^{k+1} } 
\right| \mathbf{c} = \mathbf{c}^k 
\right]
\\
&\ =\
1_A \left[
\frac{ c_2 }{ c_1 + 1 } p
+
\frac{ c_2 + 1 }{ c_1 } (1-p) \frac{ c_2 }{ c_1 + c_2 }
+
\frac{ c_2 }{ c_1 } (1-p) \left( 1 - \frac{ c_2 }{ c_1 + c_2 }
\right) 
\right]
\\
&\ =\
1_A \left[
\frac{ c_2 }{ c_1 } 
\left[
1
-
\frac{ 1 }{ c_1 + 1 } p
\right]
+
\frac{ 1 }{ c_1 } 
(1-p) \frac{ c_2 }{ c_1 + c_2 }
\right]
\\
&\ \leq\
1_A 
\frac{ c_2 }{ c_1 } 
\left[
1
-
\frac{ 1 }{ c_1 + 1 } p
+
\frac{ 1 }{ c_1 + 1 } 
(1-p) 
\right]
\\
&\ \leq\
1_A 
\frac{ c_2 }{ c_1 } 
\left[
1
-
\frac{ 1 }{ k + c_1^0 + 1 } ( 2p - 1)
\right]
\end{align*}
Using Lemma 3 and the above, the proof of statements a), b), c) and d) are 
exactly is in the proofs
of the Fundamental Theorem for Speakers and   
the Fundamental Theorem for Hearers. 
$\blacksquare$

\begin{thrm} {\bf  Form Competition Model, $p<1/2$}

Suppose $p = P( m_{\textsc{subj}} ) < .5$ in the Form
Competition production algorithm.  Then, for any start values
$c_1^0$, $c_2^0 > 0$, as the number of utterances in the
language history grows, we have as $k \to \infty$


a) the probability a speaker with message
$m_{\textsc{subj}}$ chooses form $f^u$ is always 1,

b) the probability a speaker with message $m_{\textsc{obj}}$ 
chooses form $f^u$ converges to $(1 - 2p) / ( 1 - p)$,

c) the probability a hearer interprets $f^u$ as
$m_{\textsc{subj}}$ converges to $p / ( 1 - p)$,

d) the probability a hearer interprets $f^u$ as
$m_{\textsc{obj}}$ converges to $( 1 - 2p)  / ( 1 - p)$.

All convergences are almost sure. 
\end{thrm}
{\bf Proof: } 
The 
statements follow either directly from Lemma 4 below and equations (41), (42) 
in The Emergence of Grammar Through Reinforcement 
Learning or
from using Bayes theorem as in the Fundamental Theorem for Hearers. 
$\blacksquare$

\begin{lem}
Suppose $p = P( m_{\textsc{subj}} ) < .5$ in the Form
Competition production algorithm.  Then, for any start values
$c_1^0$, $c_2^0 > 0$, as the number of utterances in the
language history grows, we have as $k \to \infty$,
$c^k_1/(c_1^k + c_2^k) \to p / ( 1- p )$, 
$c_2^k/(c_1^k + c_2^k) \to ( 1-2p ) / ( 1- p )$, almost surely. 
\end{lem}
{\bf Proof: }  We use the methods of stochastic approximation
and martingales
given in 
\cite{Pemantle2007}. 
By Lemma 3, 
both $c_1$ and $c_2$ grow without bound, hence we can 
ignore utterances $f^a$ since they do not affect $P(f^u | m_i)$.
\color{black} More precisely, we let
$N(k) = c_1^k + c_2^k$ be  
the number of successful utterances (plus starting values) 
after $k$ iterations of the production algorithm. 
We get a Markov process 
$\left\{ \left(d_1^n,d_2^n \right) \right\}_{ n =N(0)}^\infty$ 
defined by   
$\left(d_1^{N(k)},d_2^{N(k)} \right) = \left(c_1^k,c_2^k \right)$ 
for $k=0,1,\ldots$, with the following properties and where
we define $x^n$, $p_1(x^n)$ and $p_2(x^n)$.  
\begin{align*}
x^n
&\ =\
\frac{ d^n_2 }{ d^n_1 + d^n_2 }
\ \in\ (0,1),
\ \ \ \ \
\ \ \ \ \
1 - x^n
\ =\
\frac{ d^n_1 }{ d^n_1 + d^n_2 }
\\
p_1 \left( x^n \right) 
&\ =\
P\left( \left. d_1^{n+1} = d_1^n + 1 \, \right| \, d_1^n, d^n_2 \right) 
\ =\
\frac{ p }{ p + (1-p)x^n }
\\
p_2 \left( x^n \right) 
&\ =\
P\left( \left. d_2^{n+1} = d_2^n + 1 \, \right| \, d_1^n, d^n_2 \right) 
\ =\
\frac{ (1-p)x^n  }{ p + (1-p)x^n }
\end{align*}

Since $d_1^n + d_2^n =n$, the convergence properties of 
the two dimensional Markov process
$\left\{ \left( d_1^n / n , d_2^n / n \right)\right\} =
\left\{ \left( 1 - x^n  , x^n \right)\right\} $ 
can be deduced from those of the one dimensional Markov
process $\left\{ x^n \right\}$.    

Following Pemantle section 2.4, 
we define a vector valued function $F =(F_1,F_2)$, 
a payoff process $v^n$, for the payoffs at $n$, and $\xi^n$,
a martingale process with uniformly bounded second moment.
\begin{align*}
F_1\left( x^n \right)
&\ =\
p_1 \left( x^n \right)
-
\left( 1 - x^n \right), 
\ \ \ \ \ 
F_2\left( x^n \right)
\ =\
p_2 \left( x^n \right) 
-
x^{n} 
\\
v^{n+1} 
&\ =\
\left\{
\begin{array}{ll}
(1,0)
\ \ \ \ 
&\mbox{ with probability }\  p_1 \left( x^n \right) 
\\
(0,1)
\ \ \ \ 
&\mbox{ with probability }\  p_2  \left( x^n \right)
\end{array}
\right.
\\
\xi^{n+1}
&\ =\
v^{n+1}  
-
\left( p_1 \left( x^n \right), p_2 \left( x^n \right) \right), 
\ \ \ 
\mbox{E} \left[ \left. \xi^{n+1} \right| {\cal F}_n 
\right]
\ =\
0,
\ \ \ 
\mbox{E} \left[ \left. \left| \xi^{n+1} \right|^2 \right| {\cal F}_n 
\right]
\ \leq\
8
\end{align*}
Then process $\left\{ \left( d_1^n / n , d_2^n / n \right)\right\} =
\left\{ \left( 1 - x^n  , x^n \right)\right\} $ 
is a {\it stochastic approximation process} since
satisfies Pemantle's equation (2.6) 
with no remainder term and 
with $1/(n+1) $ replacing $1/n$.
\begin{align*}
\left( 1- x^{n+1},x^{n+1} \right)
-
\left( 1 - x^{n}, x^{n} \right)
&\ =\
\frac{ \left(d_1^n, d_2^n \right) + v^{n+1} }{ n + 1 }
-
\frac{ \left(d_1^n, d_2^n \right)  }{ n }
\\
&\ =\
\frac{1}{n + 1}\left(
F \left( x^n \right)  
+
\xi^{n+1}  
\right).
\end{align*}
Replacing $1/n$ with $1/(n+1) $ doesn't matter,
since it satisfies Pemantle's hypotheses (2.7), (2.8):
$\sum_n^\infty 1/(n+1) = \infty$   
and $\sum_n^\infty  1/(n+1)^2 < \infty $ .

Since $F_2$ is continuous on $[0,1]$, 
$F_2$ is bounded and has   
uniform sign on compact subsets of an open interior interval
between its zeros. 
Hence the one dimensional process $\left\{ x^n \right\}$ 
satisfies the hypotheses of Pemantle's Lemma 2.6. 
From Pemantle's Corollary 2.7, the process 
$\left\{  x^{n} \right\}$
converges almost surely to the zero set of $F_2$
consisting of $0$ and $( 1 - 2p ) / ( 1 - p ) $.
Hence the two dimensional process converges almost surely 
to the points $(1,0)$ and $( p, ( 1 - 2p ) / ( 1 - p ) )$.
Thus it suffices to show $P( x^n \to 0) =0$
to conclude that $ (d^n_1,d^n_2)/n \to (p, (1-2p)/ (1-p )$, 
almost surely.

A linear stability analysis indicates 
$(1,0)$ is unstable, since derivative $F'_2(0) > 0$, 
and  $( p, ( 1 - 2p ) / ( 1 - p ) )$ is stable,  since
$F_2'((1-2p)/(1-p)) < 0$. Using Pemantle 
Theorem 2.17 (see also 
\cite{benaim2006}
Theorem 9.1), and noting
that $\left| \xi^n \right|$ is uniformly bounded, we have 
$P( x^n \to 0) =0$. One can also use a slight generalization of 
Pemantle Theorem 2.9 to show $P( x^n \to 0) =0$.
It is easy to check that 
if $p < 1/2$ and $0< x <  ( 1 - 2p ) / ( 1 - p ) $ then $F_2(x) > 0$.
Pemantle's Theorem 2.9 holds for interior 
points of interval [0,1] whereas the
equilibrium point 0 is on the boundary. But $x^n = d_2^n / n $
is either always positive (when $x^0>0$) or always 0 
(when $x^0=0$).   Thus the proof of 2.9 
holds when checking $x > 0$ and restricting to open intervals of 
the form $(0,\delta)$, $\delta>0$ small, or 
$(0,c n^{-1/2})$ with some constant $c>0$.  

Hence, the two dimensional process
$\left\{ \left( d_1^n / n , d_2^n / n \right)\right\} =
\left\{ \left( 1 - x^n  , x^n \right)\right\} $ 
converges to $ ( p,  1 - 2p  ) / ( 1 - p ) $ and  
\begin{align*}
\frac{ \left( c_1^k , c_2^k  \right) }{  c_1^k + c_2^k  }
&\ =\
\frac{ \left( d_1^{N(k)} , d_2^{N(k)}  \right) }{  {N(k)}  }
\ \to\
\left( \frac{ p }{ 1 - p } , \frac{ 1 - 2p }{ 1 - p } \right)
\ \ \ \ \
\mbox{ as }\
k \to \infty.
\end{align*}
$\blacksquare$


\end{document}